\documentclass[twoside,11pt]{article}
\usepackage{jair, rawfonts}

\ShortHeadings{Model Reporting for Certifiable AI}
{Brajovic et al.}
\firstpageno{1}

\usepackage{acronym}
\usepackage{longtable}
\usepackage{booktabs}
\usepackage{subfig}
\usepackage{graphicx}
\usepackage[table]{xcolor}

\usepackage[natbibapa]{apacite} 
\usepackage{hyperref}

\newacro{EU}[EU]{European Union}
\newacro{GDPR}[GDPR]{General Data Protection Regulation}
\newacro{ML}[ML]{machine learning}
\newacro{MLP}[MLP]{multilayer perceptron}
\newacro{AI}[AI]{artificial intelligence}
\newacro{ISO}[ISO]{International Organization for Standardization}
\newacro{IML}[iML]{interpretable machine learning}
\newacro{PCA}[PCA]{principal component analysis}
\newacro{XAI}[xAI]{explainable artificial intelligence}
\newacro{HPO}[HPO]{Hyperparameter Optimization}

\begin{document}

\title{Model Reporting for Certifiable AI:\\ A Proposal from Merging EU Regulation into AI Development}

\author{\name Danilo Brajovic$^*$            \email danilo.brajovic@ipa.fraunhofer.de \\
       \name Philipp Wagner$^*$              \email philipp.wagner@ipa.fraunhofer.de \\
       \name Mara Kläb                       \email mara.klaeb@ipa.fraunhofer.de \\
       \name Benjamin Fresz                  \email benjamin.fresz@ipa.fraunhofer.de \\
       \name Marco F. Huber                  \email marco.huber@ieee.org\\
       \addr University of Stuttgart, Institute of Industrial Manufacturing and Management IFF,   and\\ 
                Fraunhofer Institute for Manufacturing Engineering and Automation IPA\\ 
                Department Cyber Cognitive Intelligence (CCI)\\
                Nobelstr. 12, 70569 Stuttgart, Germany
\AND
       \name Vincent Philipp Goebels$^*$    \email vincent-philipp.goebels@iao.fraunhofer.de \\
       \name Janika Kutz                    \email janika.kutz@iao.fraunhofer.de \\
       \name Jens Neuhuettler               \email jens.neuhuettler@iao.fraunhofer.de \\
       \addr  Fraunhofer Institute for Industrial Engineering IAO\\
                Bildungscampus 9, 74076 Heilbronn, Germany
\AND
       \name Niclas Renner$^*$              \email niclas.renner@iat.uni-stuttgart.de \\
       \addr University of Stuttgart, Institute of Human Factors and Technology Management IAT\\
       Nobelstr. 12, 70569 Stuttgart, Germany
\AND
       \name Martin Biller                  \email martin.biller@audi.de \\
       \addr AUDI AG,
                Data Analytics N/P1-441 \\
                Neckarsulm, Germany
}


\maketitle

\def\thefootnote{*}\footnotetext{Equal contribution. Each author provided one of the cards. First author coordinated writing.}\def\thefootnote{\arabic{footnote}}

\begin{abstract}
Despite large progress in Explainable and Safe AI, practitioners suffer from a lack of regulation and standards for AI safety. 
In this work we merge recent regulation efforts by the European Union and first proposals for AI guidelines 
with recent trends in research: data and model cards. 
We propose the use of standardized cards to document AI applications throughout the development process. 
Our main contribution is the introduction of use-case and operation cards, along with updates for data and model cards to cope with regulatory requirements.
We reference both recent research as well as the source of the regulation in our cards and provide references to additional support material and toolboxes whenever possible.
The goal is to design cards that help practitioners develop safe AI systems throughout the development process, while enabling efficient third-party auditing of AI applications, being easy to understand, and building trust in the system. Our work incorporates insights from interviews with certification experts as well as developers and individuals working with the developed AI applications.

\end{abstract}

\section{Introduction}
One key enabler for the application of \ac{AI} and \ac{ML} is the establishment of standards and regulations for practitioners to develop safe and trustworthy \ac{AI} systems. Despite several efforts by different research institutions and governmental organizations, there are  currently no established guidelines. The most influential attempt stems from the European Union with its AI Act \citep{AiAct}.  
In this paper we revisit  the state of the art in safeguarding and certifying AI applications with a focus on the the European Union. Despite this, our summary is intended to be as broad as possible. The focus on European legislation stems from the influence on regulation and standardization it has.\footnote{\href{https://time.com/6278144/european-union-artificial-intelligence-regulation/}{European Union Set to Be Trailblazer in Global Rush to Regulate Artificial Intelligence (TIME)}} 
The result is a structured guideline for reporting AI applications along four major development steps.
The objective of our guideline is to provide comprehensive support during the development process. It encompasses a synthesis of requirements derived from both non-technical users and prominent certification and standardization bodies. Our approach involves the inclusion of robust references to relevant support materials and toolboxes, emphasizing the transparency and traceability of each requirement's source. For this purpose, we conduct interviews with standardization experts, developers and consumers of the final AI product and incorporate their feedback into our proposal. Furthermore, our framework serves as a solid foundation for prospective system certification. By adhering to our guidance, developers can ensure they are adequately equipped to meet forthcoming legal obligations within their domain.
The key contributions of our work is fourfold: 
\begin{itemize}
    \item Our guideline is meant to be used during the development process. 
    \item We combine requirements from non-technical users as well as certification and standardization bodies.
    \item We include references to support material and toolboxes and link the source of the requirement whenever possible.
    \item We lay the foundations for a future certification of the system. With our framework, developers should be well-positioned when legal requirements  come into force.
\end{itemize}

\subsection{Safeguarding AI, Law, Certification, Standardization, and Audits}
Before going into the specifics of AI regulation, we first give a short summary of relevant terms in EU regulations.
The term \emph{audit} describes the process of testing whether a product, service, or process meets certain requirements.
The result of such a process can be a \emph{certification}, the assurance by a third-party auditor that given requirements are fulfilled.
These requirements are often listed in \emph{harmonized standards}, which are created by organisations like CEN, CENELEC, or ETSI, following a request from the European Commission.
These harmonized standards provide the technical details necessary for companies to be able to ensure compliance with regulation and for auditors to have clear specifications to check against.
The process of creating such harmonized standards for the EU AI Act is currently running in parallel to the legislation procedure.
As a baseline for the standards harmonized with the EU AI Act, standards provided by ISO or other standardization bodies can be considered.
Conformity assessment bodies can then utilize these standards to showcase the compliance of products, services, or processes with applicable EU legislation \citep{EU-harmonised-standard}.
These audits can also be carried out on a voluntary basis for products that do not necessarily require them, for instance for marketing purposes.
In certain product categories like medical devices, third-party conformity assessments are mandatory to ensure compliance with technical regulations before entering specific markets \citep{EU-conformity-assessment}.
Apart from such certificates, so called \emph{guidelines} have no legally binding effect.

Finally, developing safe and responsible AI also involves governance structures within the company to manage data development activities and risks. However, this is beyond the scope of this paper, in which we focus on reporting around the development process. 

\subsubsection{Challenges in AI Certification}
Despite being a  topic of active research for several year, there is a lack of experience and methods for securing and certifying \acsu{ML} applications.
Apart from the fast progress in  \acsu{ML} around research and development, the challenges are rooted in the fundamental programming differences between classic software and \acsu{ML}. 
Classic software components are usually a combination of manually defined functions, the logic within these functions being explicitly specified by a human. In contrast, the core of \acsu{ML} is to learn from data while the user only specifies the model framework. Often, the result is a highly complex black-box that returns results without explaining the underlying decision-making process. 
The main challenge of certifying such a system is validating a black-box model representing a highly complex function.  Furthermore, an \acsu{ML}  algorithm  always entails risks related to data, e.g., the risk of a selection bias \citep {Tidjon2022_ChallengesAI}. 
Hence, within the certification process, not only does the proper functionality of the model need to be verified but also the representability of the data for training and testing. This is a challenging task, because the reason for developing \acsu{ML} often is that the underlying problem and cause-effect relations are too complex to be understood, formulated and solved analytically and manually.

\subsection{The State of \ac{AI} Certification}
There are several attempts by different organizations to standardize and secure the application of \ac{AI}. Apart from several ISO standards which are currently under development (e.g. \cite{ISO42001} - Artificial intelligence - Management system)  the European \ac{AI} Act \citep{AiAct} and the AI Assessment Catalog by Fraunhofer IAIS \citep{poretschkin2021_IAIS_Leitfaden} emerged during our discussions with organizations working on AI standardization and as such form a baseline for this work. They are described in more detail in this section.

\subsection{Non-academic Activities}
\subsubsection{The European \ac{AI} Act}
The most impactful work comes from the \ac{EU} with the proposal of the \ac{AI} Act~\citep{AiAct}. The first draft was proposed in April 2021 by the European Commission and two comments were recommended in December 2022 from the \cite{AiAct_Ratskompromiss} and in April 2023 by the \cite{AiAct_Parlamentsvorschlag}. 
The initial draft already created a heated debate leading to fears that it might over-regulate research and development of \ac{AI} in the \ac{EU} and that the definition of \ac{AI} was too broad. However, this debate is beyond the scope of this paper. We are interested in the requirements posed on those systems. Although some are still subject to change, many requirements have already emerged. 

In general, the \ac{AI} act distinguishes three risk classes for \ac{AI}: forbidden applications, high-risk applications and uncritical applications. Furthermore, there might be some obligations of transparency even for uncritical applications. For example, that end-users must be informed that they are interacting with an \ac{AI} system. There are no mandatory obligations for uncritical applications---although developers of such \ac{AI} systems shall still be motivated to follow the proposed principles. 

Whether an application falls into the high-risk category is risk-dependent, and a few high-risk applications are named in the \ac{AI} Act's appendix. Some of them include: biometric identification, operation of critical infrastructure, education, employment \& workers management, essential private services \& public services, law  enforcement, migration \& asylum, and administration of democratic processes. The requirements for such systems are listed in Chapter~2 and include 
\begin{itemize}
    \item a risk management system (Article 9)%
    \item  data and data governance (Article 10)%
    \item  technical documentation (Article 11)%
    \item record-keeping (Article 12)%
    \item  transparency and provision of information to users (Article 13)%
    \item  human oversight (Article 14)%
    \item  accuracy, robustness and cybersecurity (Article 15). 
\end{itemize}

A further distinction is based on whether an AI system is considered as \emph{general purpose AI} (\emph{GPAI}) or \emph{foundation model}. Although the term GPAI lacks a proper and widely accepted definition, GPAI should refer to systems such as image and speech recognition that  can by applied to a broad variety of tasks. Foundation models, on the other hand, will most likely be distinguished by the training data and the generality of the output. This should specifically target systems such as large language models like ChatGPT and pose stricter regulation on them while GPAI systems must only comply with the regulations in the Articles 9--15 if they are expected to be used in high-risk scenarios. The definitions for GPAI and foundation models (and others) are provided in Article 3. Finally, Chapter 3 lists further obligations for providers of high-risk AI systems and will most likely contain the requirements for foundation models once the AI Act is final.
At the moment, it is highly likely that a self-assessment will be possible for most applications. This means that companies can audit their applications by themselves and then apply the \href{https://en.wikipedia.org/wiki/CE_marking}{CE-mark} to their product. Third party assessments will only be mandatory for certain highly critical domains such as medical technology. 
Overall, the requirements are on a very abstract level and still subject to change. In particular, the act does not contain a specific starting points for practical implementation in companies because the baseline standards that will provide the technical details are still being developed.  

\subsubsection{\ac{AI} Assessment Catalog}
The \emph{\ac{AI} Assessment Catalog} is the most comprehensive and technical guideline with over 100 pages of content. It gives authors and practitioners a structured tool to develop and asses trustworthy \ac{AI} systems \citep{poretschkin2021_IAIS_Leitfaden}. The catalog considers the whole \ac{AI} application, i.e., the data and the AI components together with surrounding non-AI software components, and identifies six risk dimensions that need to be considered: fairness, autonomy and control, transparency, reliability, safety, and privacy. For each of these six dimensions there are guidelines for risk analysis and appropriate actions. The actions for each risk belong to one of the four groups: data, \ac{AI} components, embedding, or operations. Thus, if developers want to check whether they have addressed all necessary actions for data they have to consult all six risk dimensions. To the best of our knowledge, the AI Assessment Catalog is the most comprehensive guideline for trustworthy AI. However, due to its size the catalog can be cumbersome for practitioners to use. Finally, although being developed by a German organization, the guideline is not specific to Germany nor the \ac{EU} but combines elements from different areas of research and development.  

\subsection{Research Activities}
In addition to government and industry activities, there is research work on model and data reporting with cards that finds application especially among larger corporations.\footnote{\href{https://ai.googleblog.com/2022/11/the-data-cards-playbook-toolkit-for.html}{The Data Cards Playbook: A Toolkit for Transparency in Dataset Documentation}} \footnote{\href{https://ai.googleblog.com/2020/07/introducing-model-card-toolkit-for.html}{Introducing the Model Card Toolkit for Easier Model Transparency Reporting}} Our own work builds on documentation along these cards. 

\subsubsection{Model Cards}
\cite{Mitchell2019_ModelCards}  propose model cards as a standardized way to document \ac{ML} models, their performance metrics and characteristics. The work focuses primarily on ethics and fairness, citing that AI models in recent history have often erred more on social groups that have historically been marginalized in the US. All involved stakeholders benefit from a standardized use of model cards. The authors list nine sections for model cards:  model details,  intended use, factors, metrics,  evaluation data,  training data,  quantitative analyses,  ethical considerations, and caveats and recommendations.
In our proposal we divide model and data cards and therefore exclude the section on data. 

\subsubsection{Data Cards}
Data cards are similar to model cards \citep{Pushkarna2022_DataCards}. Their goal is to report several standard facts about a dataset in a simple and standardized form. The authors list several requirements for their cards. In particular, they should be standardized in order to be comparable, should be created at the same time as the dataset, written in a simple language such that users with a non-technical background can understand them, and, finally, report \emph{known unknowns} such as possible shortcomings or uncertainties of the dataset. Apart from that, the authors name 31 content themes to describe a dataset. They are listed in Appendix \ref{app:datasheets_content_themes}. 

Another related work is data sheets for datasets \citep{Gebru2021_DataSheets}. The authors propose a similar set of questions for dataset creators during the collection phase that can be grouped into the following categories: (i) \emph{motivation} for creating the dataset, (ii) \emph{composition} of the data, (iii) \emph{collection process}, (iv)  \emph{preprocessing/cleaning/labeling} of the data, (v) \emph{Uses} (describing the contexts in which the dataset was already used), (vi) the \emph{distribution} of the data, and (vii) the future \emph{maintenance}. The intended audience are both the dataset creators and the consumers of the final dataset.

\subsubsection{Foundation Models \& AI Act}
Recently, \cite{bommasani2023eu-ai-act} assessed popular foundation models like GPT-4, LLaMA and Luminous for their compliance with the AI Act. They derived and summarized 12 requirements grouped into data, compute, model and deployment from the AI Act. Our own approach is similar to their work but is not focused on foundation models. However, many of the requirements for foundation models also apply to high-risk AI systems. A summary of the requirements can be found in Table \ref{tab:stanford_llm_audit}.

\renewcommand\arraystretch{1.3}
\begin{table}
    \centering
    \begin{longtable}{p{0.15\linewidth}|p{0.6\linewidth}p{0.15\linewidth}}  
        \hline
         \textbf{Category}  & \textbf{Requirement}                                              & \textbf{Source}  \\
         \hline \hline
         \multirow{3}{=}{Data}
                            & Describe data sources                                             &  Annex VIII \\
                            & Apply data governance                                             &  Art. 28 b)\\
                            & Disclose copyrighted data                                         &  Art 28 b) \\
         \hline
         \multirow{2}{=}{Compute}
                            & Disclose compute (computer power, training time)                  &  Annex VIII \\
                            & Measure energy consumption                                        &  Art 28 b) \\
         \hline 
         \multirow{6}{=}{Model}
                            & Describe capabilities and limitations of model                    &  Annex VIII \\
                            & Describe foreseeable risks and associated mitigations             &  Annex VIII, Art. 28 b) \\
                            & Benchmark model on standard benchmarks                            &  Annex VIII, Art. 28 b)  \\
                            & Report results of internal and external tests                     &  Annex VIII, Art. 28 b)  \\
         \hline 
         \multirow{4}{=}{Deployment}
                            & Disclose content generated from the model                         &  Recital 60 g \\
                            & Disclose EU member states where model is on market                &  Annex VIII \\
                            & Provide documentation for downstream compliance with AI Act       &  Art 28 b) \\
         \hline 
    \end{longtable}
    \caption{Requirements for foundation models from \cite{bommasani2023eu-ai-act}.}
    \label{tab:stanford_llm_audit}
\end{table}

\subsubsection{Other Activities}
\label{sec:other_activities}
Besides the European \ac{AI} Act there are several other regulatory attempts. 
The EU has been actively involved in standardization activities related to \ac{AI} and plans to implement liability rules on \ac{AI} to renew the existing rules on product liability. The new rules will hold manufacturers and importers of AI-powered products accountable for any defects or malfunctions that may cause harm to consumers \citep{EULiability}. Additionally, there are a number of ISO standards either in active development or already in use that target \ac{AI} systems in particular. The ISO/IEC JTC 1/SC 42 is the technical committee within the \ac{ISO} that is responsible for developing standards for \ac{AI}. Currently, there are 57 standards planned, some of which are already published. An overview can be found on the web page of the committee.\footnote{\href{https://www.iso.org/committee/6794475/x/catalogue/p/1/u/1/w/0/d/0}{Standards by ISO/IEC JTC 1/SC 42 : Artificial intelligence}}
Among the most important ones are: the \cite{ISO5259} series on data quality, \cite{ISO5338} on AI system life cycle processes, \cite{ISO8183} on a data life cycle framework, \cite{ISO23894} on risk management for AI systems and, \cite{ISO42001} for AI management systems. The last one is not addressed in this work because it deals mostly with governance, which we do not cover here. These norms emerged from discussions with certification bodies that we held in the course of our project.

Another recent and closely related work introduces a similar guideline to ours \citep{Wittenbrink2022_LeitfadenQualitatsmanagementKI}. It breaks down the AI development process into four steps and provides requirements as well as further instructions such as relevant norms for each requirement. The main difference to our work is, that the development process is broken down into characterisation, design, development, and deployment, whereas we follow an approach oriented on existing work on model and data cards. 

On the governmental side, China and the US are among the most important other players in regulating AI. The US approach significantly differs from the European one and is (currently) based on voluntary sector-specific measures. The most important non sector-specific work is the AI Risk Management Framework developed by the National Institute of Standards and Technology \emph{(NIST)} \citep{NISTAIRMF_2023}. It defines seven key characteristics for trustworthy AI that include (i) reliability, (ii) safety, (iii) security and resiliency, (iv) accountability and transparency,  (v) explainability and interpretability, (vi) privacy, and, (vii)  fairness. The companion \href{https://airc.nist.gov/AI_RMF_Knowledge_Base/Playbook}{NIST AI RMF Playbook} names four pillars to mitigate these risks: (i) map: risks are identified, (ii) measure: the identified risks are analyzed and tracked, (iii) manage: risks are prioritized and acted upon and, finally, (iv) govern: a culture of risk management and governance is present. The key characteristics of trustworthy AI are well aligned to those defined in the AI Act.\footnote{\href{https://www.nist.gov/system/files/documents/2023/01/26/crosswalk_AI_RMF_1_0_OECD_EO_AIA_BoR.pdf}{An illustration of how NIST AI RMF trustworthiness characteristics
relate to the OECD Recommendation on AI, Proposed EU AI Act, Executive Order 13960, and Blueprint for an AI Bill of Rights}} Furthermore, NIST is working to 
 align the framework with international standards that are also relevant for the AI Act.\footnote{\url{https://www.nist.gov/artificial-intelligence/technical-ai-standards}}

China, on the other side, follows a vertical approach\footnote{\href{https://carnegieendowment.org/2023/02/14/lessons-from-world-s-two-experiments-in-ai-governance-pub-89035}{Lessons From the World’s Two Experiments in AI Governance (carnegieendowment.org/)}} that targets each AI applications with a specific regulation and aims to be a leader in AI standardization \citep{Huw_2021_ChinaAIRegulation}. For example, one regulation specifically targets services that generate text, images or videos. Overall, although the high level pillars are expected to be similar to the ones in western countries and include that AI should benefit human welfare, China's regulation is expected to be focused more around group relations than individual rights \citep{Huw_2021_ChinaAIRegulation}. However, this is beyond the scope of our paper and Chinese regulation is not addressed in our framework.

\subsection{Contributions and Methods}
We merge contributions of all of the previous works into a single framework for reporting AI applications. As mentioned, this work focuses on the documentation of the AI application and does not consider governance  within a company. 
Our guideline follows the development process of an AI system with the goal to support developers during various model development steps in creating trustworthy AI systems from the beginning and positioning them well when legal requirements come into force. Apart from this, we cover the same risk dimensions as previous approaches and reference the AI Act and other guidelines whenever possible. The AI development cycle is broken down into four steps: (i) the use-case definition, (ii) data collection, (iii) model development, and finally,  (iv) model operation. We use four cards as method of reporting in order to reduce the number of resulting documents so  the hurdle to access them is as low as possible while the different topics remain separated. In the coarse of the project we furthermore evaluated the approach on three use-cases and integrated both feedback from developers and other affected individuals. Finally, we also held interviews with several certification bodies to get an understanding about the most important existing documents and guidelines for our work.

\subsubsection{Motivation for Cards}
Apart from discussions with certification experts preparing for the certification of AI systems, we also polled employees being affected by the introduction 
of AI systems. Our goal was to assure that our proposal would both satisfy the requirements of possible auditors and builds trust by the individuals being affected by the AI solution. 
The affected individuals mainly wished to be included in the development process in order to convince themselves about the functionally of the AI system. Furthermore, the system should be transparently visualised and the output and reasons for the decision should be comprehensibly explained. 
The reason why we choose the format of cards that follow the development process instead of following risk dimensions (in contrast to, for instance, \cite{poretschkin2021_IAIS_Leitfaden}) is twofold: firstly, for developers it is easier to follow the development process and for domain experts and other employees the resulting number of documents is smaller and as a result easier to access.
Secondly, although AI development is often an agile process, the cards can still be utilized by updating them when the dataset changes or the model is updated. They also integrate well into a classical v-model development, which is subject to future work.

\newpage
\section{Proposed Approach}
In this section, we propose cards supporting the four steps: use case definition, data collection, model development, and model operation. For each card, we briefly cover background information before presenting our proposal and include references to further literature whenever possible. Furthermore, each card contains an \emph{involvement of affected individuals} section as a reminder to include people that might be affected by the AI system whenever feasible.
In Appendix \ref{app:audit} we provide an example on a toy use case. As mentioned before, we focus on reporting around the AI application and do not cover governance. 

\subsection{Use Case}

In line with \cite{Raji2020_Accountability} and \cite{poretschkin2021_IAIS_Leitfaden} the first step in our framework is a summary of the use case. This offers the auditing authority a brief overview about the task at hand and the AI application. Furthermore, the document can be used during the AI development phase to ensure compliance as well as a trustworthy and robust AI system. A general understanding of the use case is necessary to perform a proper risk assessment afterwards. Therefore, a use case summary should be formulated that includes a description of the status before applying an AI solution, a description of relevant sub-components as well as a summary of the proposed AI solution. This approach is especially useful for the documentation of high-risk applications as defined in the AI Act \citep{AiAct}. A general summary should also include a contact person and information about involved groups inside an organisation to include them in relevant parts during the development phase. New in our framework is the justification of why a non-AI approach is not sufficient in the context of the application. Although this requirement is not explicitly stated in the AI Act, it emerged in discussions with several certification bodies. For developers this means that a simpler model (or maybe non-AI) solution should be prioritized whenever there is no sound justification for the use of a more complex one. Finally, two new aspects in our approach are (i) reviewing and documentation of prior incidents in similar use cases and (ii) the distinction whether a general purpose AI system or a foundation model is expected to be used or developed. 

\subsubsection{Risk} 
AI can only be deployed if the associated risks are kept on an acceptable level, which is comparable to already existing products or services. To guarantee this, a proper risk assessment must be performed. Due to the nature of AI systems, e.g., the training on noisy and changing data or the black-box characteristics, classical methods for risk assessment cannot be applied in most cases \citep{Aria2020_VS, Siebert2022_Quality}. In literature, there are several schemes for qualitative and quantitative risk assessment. In recent years, the AI community proposed several ways to categorize risks in different dimensions \citep{poretschkin2021_IAIS_Leitfaden, Ashmore2021_Assuring, Piorkowski2022_Risk}. \pagebreak 
One approach, accepted by many scientists and organisations, is the division according to the High-Level Expert Group on AI \citep{EU2019_Ethic}. 
They propose the seven key dimensions of risk: 
\emph{
    \begin{enumerate}
        \item human agency/oversight, 
        \item  technical robustness/safety, 
        \item  privacy/data governance, 
        \item  transparency, 
        \item  fairness, 
        \item  environmental/societal well-being, and 
        \item  accountability. 
    \end{enumerate}
}
As described in Section \ref{sec:other_activities}, these dimensions are well-aligned to other works. Based on every dimension one can derive possible risks for one specific use case and AI application. Here, the risk consists of the probability of an undesirable event and the impact such an event has. In recent years, many more methods have been developed to accurately quantify such probabilities, for example, in the field of adversarial robustness \citep{Szegedy2013_Adversarial, Murakonda2020_Privacy}, fairness \citep{Barocas2016_BigDD, Bellamy2018_360, bird2020_fairlearn}, or privacy \citep{Shokri2016_privacy, Fredrikson2015_Inversion}. In the context of (functional) safety, a quantitative assessment of the risk is almost always mandatory, especially in safety critical applications such as robotics, manufacturing, or autonomous driving \citep{El-Shamouty2022_GLIR,Ashmore2021_Assuring, Salay2017_26262}. There are also more qualitative approaches to classify the risk corresponding to one dimension. The \ac{EU} separates between forbidden applications, high risk applications and uncritical applications based on the application domain \citep{AiAct}. 
A standardized way of performing risk assessment for AI is given in ISO/IEC 23894 which uses the existing risk management standard ISO 31000 as a reference point. It also offers concrete examples in the context of correct risk management integration during and after the development.

Our proposal of a use case documentation is shown below. In accordance with existing works on model cards \citep{Mitchell2019_ModelCards} and data cards \citep{Pushkarna2022_DataCards} we call it a \emph{Use Case Card}. As in the other cards for almost every aspect there exist one or multiple references to the AI Act. After a use case summary and a brief formulation of the problem and solution, we suggest to perform a risk assessment based on the dimensions mentioned above. The risks has to be stated associated with the corresponding dimension. Note that in this approach first a qualitative risk assessment is performed to create awareness of the critical aspects of the AI system. Later on, in the Sections \ref{subsec:data}--\ref{subsec:monitoring}, the goal is to quantify and minimize those risks using specific methods and other measures.

\newpage

\label{tab:uc}
{\small
    \renewcommand\arraystretch{1.3}
    \begin{longtable}{p{0.15\linewidth}|p{0.35\linewidth}p{0.15\linewidth}p{0.35\linewidth}} 
    \toprule
    \multicolumn{4}{c}{\bf{Use Case Card}}\\
    \toprule
    \bf{Step}        & \bf{Requirement}      & \bf{AI Act}& \bf{References} \\
    \hline
    \hline
    \multirow{12}{=}{General}
                        & Name contact person & Art. 16, Art. 24 & \\\cline{2-4}
                        & List groups of people involved & & \\\cline{2-4}
                        & Summarize the use case shortly & Art. 11 & \cite{poretschkin2021_IAIS_Leitfaden}\\\cline{2-4}
                        & Describe the status quo & & \\\cline{2-4}
                        & Describe the planed interaction of sub-components (e.g., different software modules or hardware and software) & Art. 18 & \\\cline{2-4}
                        & Provide a short solution summary & & \cite{Mitchell2019_ModelCards}\\\cline{2-4}      
                        & Review and document past incidents in similar use cases & & \href{https://incidentdatabase.ai/}{AI Incident Database}\\\cline{2-4}
                        & Review available tools & & \href{https://oecd.ai/en/catalogue/overview}{Catalogue of Tools \& Metrics for Trustworthy AI}\\

    \specialrule{.075em}{0em}{0em}
    \multirow{5}{=}{Problem definition}
                        & Clearly describe the learning problem & Art. 11 & \\\cline{2-4}
                        & Describe the disadvantages of the current approach & & \\\cline{2-4}
                        & Argue why classical approaches are not sufficient & & \\
                        
    \specialrule{.075em}{0em}{0em}
    \multirow{7}{=}{Solution approach}
                        & Describe the integration into the current workflow & Art. 11 & \\\cline{2-4}
                        & Formulate the learning problem & & \\\cline{2-4}
                        & Formulate the KPIs for go-live & & \\\cline{2-4}
                        & Document if a foundation model will be used or developed & Art. 3, Art 28 b) & \\\cline{2-4}
                        & Provide a short data description & Art. 10 & 
                        \cite{Gebru2021_DataSheets}\\
                        
    \specialrule{.075em}{0em}{0em}
    \cellcolor[HTML]{d3d3d3}& \cellcolor[HTML]{d3d3d3}\textbf{Risk Assessment} &\cellcolor[HTML]{d3d3d3} Art. 9, Art. 19 &\cellcolor[HTML]{d3d3d3} \\
    \hline
    \hline
    \multirow{2}{=}{Risk Class}          
                        & Categorize the application into a risk class according to the AI Act & Art. 5, Art. 6 & \href{https://blog.burges-salmon.com/post/102i9my/navigating-the-eu-ai-act-flowchart}{Navigating AI Act Flow Chart} \href{https://appliedaiinitiative.notion.site/Risk-Classification-Database-2b58830bb7f54c9d8c869d37bdb27709}{Risk-Classification DB}\\
                        
    \specialrule{.075em}{0em}{0em}
    \multirow{2}{=}{Human agency and oversight} 
                        & Rate the level of autonomy & Art. 14 \\
                                                &                   &\\
    \specialrule{.075em}{0em}{0em}
    \multirow{8}{=}{Technical robustness and safety}
                        & Evaluate the danger to life and health & Art. 5 (1) a), Art. 14 & \cite{winter2021_TueV_Austria} \\\cline{2-4}
                        & Identify possibilities of non-compliance & Art. 15, Art. 19 & \cite{johnson2020understanding}\\\cline{2-4}
                        & Identify customer relevant malfunction & & \\\cline{2-4}
                        \pagebreak
                        \cline{2-4}
                        & Identify internal malfunction & & \\\cline{2-4}
                        & Evaluate cybersecurity risks & Art. 15, Art. 42 & \\
                        
    \specialrule{.075em}{0em}{0em}
    \multirow{5}{=}{Privacy and data governance}
                        & List risks connected with customer data & Art. 10 & \cite{Cristofaro2020_Privacy}\\\cline{2-4}
                        & List risks connected with employee data & & \\\cline{2-4}
                        & List risks connected with company data & & \\
                        
    \specialrule{.075em}{0em}{0em}
    \multirow{4}{=}{Transparency}
                        & Evaluate effects of incomprehensible decisions or the use of a black box model & Art. 13 & \cite{larsson2020transparency, ehsan2021expanding, felzmann2019transparency, rudin2019stop, castelvecchi2016can} \\\cline{2-4}
                        
    \specialrule{.075em}{0em}{0em}
    \multirow{6}{=}{Diversity, non-discrimination and fairness}
                        & Check for possible manipulation of groups of people & & \cite{ISO12791_Fairness, ashton2022problem, botha2020fake}  \\\cline{2-4}         
                        & Check for discrimination of groups of people regarding sensitive attributes & Art. 5 (1) b) and c) &  Section\ref{app:protected_attrs}, \cite{ISO12791_Fairness, Handbook2018_NonDiscrimination, Oneto_2020}\\\cline{2-4}
                        
    \specialrule{.075em}{0em}{0em}
    \multirow{9}{=}{Societal and environmental well being}
                        & List possible dangers to the environment & & \cite{wu2022sustainable} \\\cline{2-4}
                        & Consider ethical aspects & & \cite{dubber2020oxford, muller2020ethics, bostrom2018ethics, hagendorff2020ethics, EU2019_Ethic} \\\cline{2-4}
                        & Evaluate effects on corporate actions & & \cite{glikson2020human}  \\\cline{2-4}
                        & Identify impact on the staff & & \cite{malik2022impact} \\
                        
    \specialrule{.075em}{0em}{0em}
    \multirow{3}{=}{Accountability}
                        & Estimate the financial damage on failure & Art.17 & \cite{gualdi2021artificial} \\\cline{2-4}
                        & Estimate the image damage on failure & & \\
                        
    \specialrule{.075em}{0em}{0em}
    \multirow{3}{=}{Norms}
                        & List relevant norms in the context of the application (e.g., automotive safety norms in an automotive use-case) & Art. 9 (3) & \cite{gasser2020role} \href{https://aistandardshub.org/ai-standards-search/}{AI Standards Hub}\\
                        
    \specialrule{.075em}{0em}{0em}
    \multirow{2}{=}{Involvement of Individuals}    
                        & If feasible, describe involvement of affected individuals  & \\
    \bottomrule
\end{longtable}
}

\newpage

\subsection{Data}
\label{subsec:data}
AI applications stand and fall with the training data. Despite that, there has only recently been a paradigm shift towards the datasets due to the data-centric AI perspective. The AI Act requires in Article 10 (3) that \emph{training, validation, and testing datasets shall be relevant, representative, free of errors and complete}. What this means in practice, however, is often unclear. Our extension of a data card builds on previous work on data documentation but is extended with further references, especially those addressing regulation, whenever possible \citep{Pushkarna2022_DataCards, Gebru2021_DataSheets}. From the standardization side, the \cite{ISO5259} series deals with data quality for AI systems, \cite{ISO8183} with the data life cycle framework for AI systems and \cite{ISO25012} describes a general data quality model. The first two are still under development but are worth reviewing once they are final. Before introducing the data card, we discuss open questions that we think are not covered sufficiently in the state of the art.

\subsubsection{Related Research} 
We start by summarizing research directions that might be worth considering before starting data collecting.
Two related active areas of research are \emph{active learning} \citep{Settles2010_ActiveLearningSurvey} and \emph{core sets} \citep{Feldman2020_CoreSetsSurvey}. In active learning, the algorithm is given access to a small labeled dataset $D$ and a large unlabeled dataset $U$. The algorithm can then query an oracle (e.g., human labeler) for labels up to some budget $b$. The goal is to optimize model performance within the  limit of the budget. A core set is a small fraction $S\in D$ of the original dataset $D$ such that the learning algorithm achieves a similar performance on this subset as if it was trained on the entire dataset.

Recently, the concept of memorization or uniqueness of data points was introduced \citep{Feldman2020_LongTail, Jiang2020_CharacterizingRegularities}.  A data instance $i$ from the training set is called unique if removing it from the training set reduces the probability for $i$ to be classified correctly by the model. \cite{Feldman2020_LongTail} hypothesizes that such rare points are important for the generalization of models. I.e., a single rare point such as an image of a red car could determine whether this concept is correctly classified by the model in the real world or not. 
\citet{Jiang2020_CharacterizingRegularities} use it as a measure to categorize the structure of a dataset and show that mislabeled points are harder to memorize. Unfortunately, estimating these instances is computationally expensive. For each data instance at least two models need to be trained; one with and the other without the instance. However, there might be occasions when it could become useful, for example, if a specific point shall be inspected. 
Finally, \citet{Meding2021_DDD} describe dichotomous data-difficulty and show that many datasets such as ImageNet suffer from imbalanced data difficulty. There are many data points in the test set that are never classified correctly (called \emph{impossible}) and many that are always classified correctly (called \emph{trivial}). They show that models can better be compared on the remaining points.  A similar conclusion can be drawn for label errors. \cite{Northcutt2021_LabelErrors} show that larger models tend to be favored on datasets with label errors while smaller models might actually outperform them when evaluated on a dataset without errors.

These research fields and recent advances are related to data collection and quality and performing a literature review before starting to collect data can help to make the collection more effective.

\subsubsection{Data Collection}
\label{subsec:data_collection}
After a literature review, the first step is to gather data. Apart from collecting metadata and protected attributes, we focus on suggestions around dataset size and coverage. It is worth mentioning that, to the best of our knowledge, there are currently no established practices in that regard and the upcoming ISO standards on data \cite{ISO5259} will most likely not include detailed requirements. Hence, the following suggestions should be seen as points worth to consider rather than being mandatory. 

\paragraph{Metadata and Leakage} Although being well-understood, leakage remains a central problem in many datasets \citep{Kapoor22_LeakageAndReproducibility}. A famous example is CheXNet for pneumonia detection on x-ray images \citep{Rajpurkar2017_CheXNet}, where the authors performed a na\"ive train-test split with overlapping x-ray images of the same patients in training and test sets in the first version of the paper. As such, the performance of the model on real data was overestimated. 
When creating train and test splits, such leaks can be avoided by collecting metadata and using it for stratification or to generate data splits of different difficulty.  In the ChexNet example, the metadata could be an anonymous ID for each patient and an easy test set would contain data from patients already represented in the train data wheres a hard test set would only contain data of unseen patients. This information can help to estimate the generalization capabilities of a model which can differ between applications. If the goal is to predict re-occurrence of a disease, performance on the easy test set could be the primary metric while performance on the hard test set would be of interest if the goal is to detect pneumonia in new patients. In general, it is advisable to collect whatever metadata is easy to obtain and compatible with data protection law because it is often impossible to backtrack once the dataset is final. This especially includes protected attributes.  

\hspace*{10mm}\subparagraph{Protected Attributes}
\label{app:protected_attrs}
Usually, the protected attributes include sex, race, skin colour, ethnic or social origin, 
genetic features, language, religion or belief, political or any other opinion, membership of a national minority, property, birth, disability, age or sexual orientation and can be found, for instance, in Art. 21 of the Charter of Fundamental Rights of the European Union \citep{CharterFundamentalRightsEU}. According to the EU handbook on non-discrimination law, 
it might differ between use-cases whether an attribute is considered as protected \citep{Handbook2018_NonDiscrimination}. Furthermore, pregnancy can be considered as protected in some cases. Access to such information is crucial when detecting biases but there may also be cases when other laws prohibit the use of such attributes \citep{van_Bekkum_2023_DoesGDPRNeedException}. 

\begin{figure}
    \centering
    \subfloat[{\scriptsize Model Performance for Train Fraction}]{{
        \includegraphics[width=0.45\textwidth]{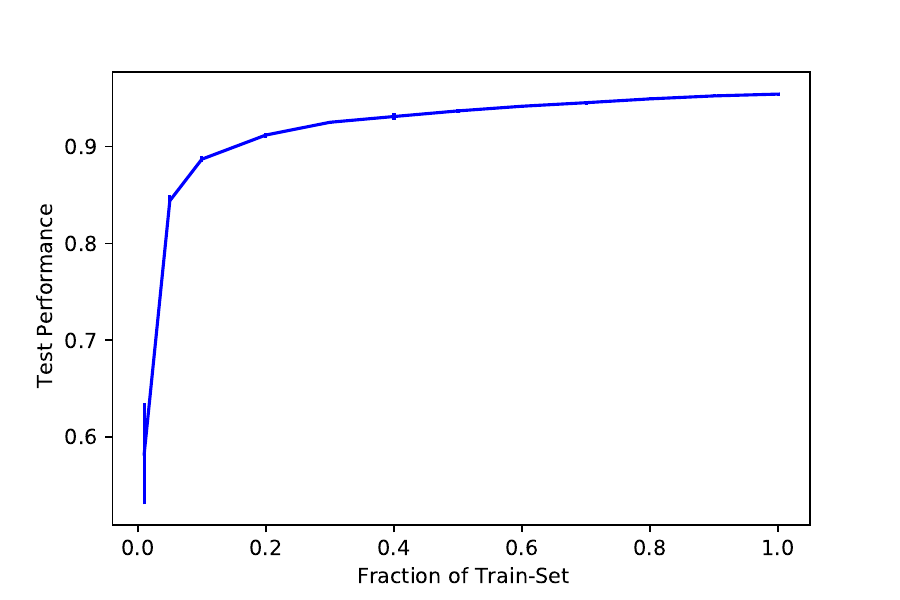}
    }}
    \subfloat[{\scriptsize Test Variance for Test Fraction}]{{
        \includegraphics[width=0.45\textwidth]{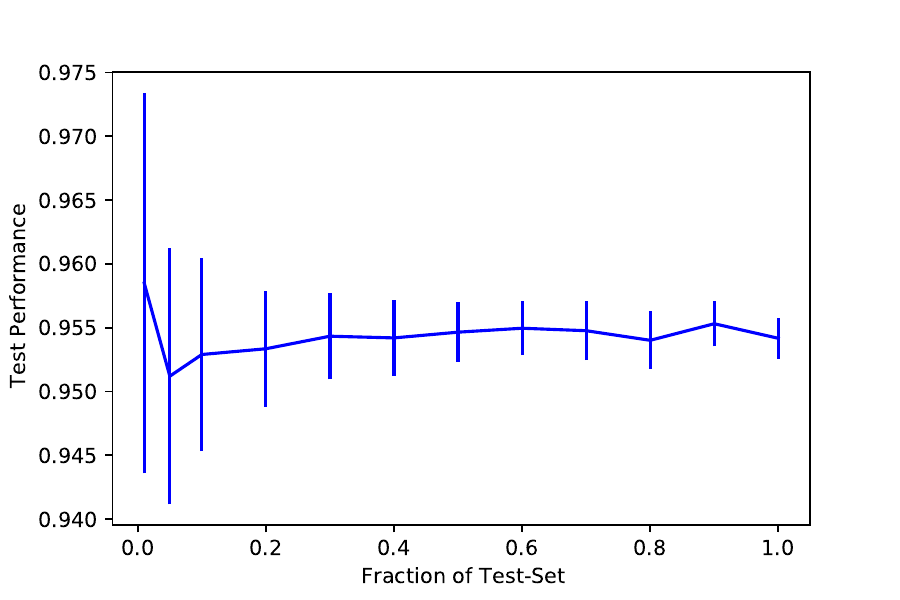}
    }}
    
    \caption{(a) Model performance of a \ac{MLP} trained on the MNIST dataset with increasing fraction of original train data. The improvement of adding data points behaves logarithmically. 
    (b) Variance over 20 runs of training an \ac{MLP} on the entire MNIST training set and evaluating the performance on fractions of the test set. With increasing size of the test set, the variance decreases.}
    \label{fig:performance_train_fraction}
\end{figure}

\paragraph{Dataset Size}
 One problem when collecting data is knowing when a sufficient amount was acquired. Despite the standard perspective that ``there's no data like more data'', according to IBM's Robert Mercer, collecting and annotating data is both expensive and not always feasible.  For example, in medicine it is almost impossible to increase the number of samples of rare diseases. Data augmentation is one solution, but in this section we want to focus on the size of the raw dataset.  A rule of thumb in computer vision suggests to use around 1.000 images per class for image classification tasks as a good starting point. The one-in-ten rule\footnote{\url{https://en.wikipedia.org/wiki/One_in_ten_rule}} suggests that per model-parameter ten data points are necessary. One way to estimate the necessary amount of data is by tracking model improvements when sequentially adding larger fractions of the training data. It is well known that the resulting performance plot behaves logarithmically \citep{Figueroa2012_PredictingSampleSize, Viering2021_LearningCurveShape}. Hence, the largest improvements are obtained early on. As shown in Figure \ref{fig:performance_train_fraction}(a), with about 50\% of the training data, the performance is already above 90\% and doubling the amount of data increases the performance by less than 5\%. In practice, one approach could be to start with a small test set for data collection and utilize it to estimate when a reasonable amount of data will be gathered. Once the data has been acquired this test set may be added to the train data or discarded and a new test set is created.

\paragraph{Data Coverage}
Apart from the size of the dataset, covering all possible scenarios is important. Recent research has shown that many data points are redundant   \citep{Toneva2018_catastrophicForgetting, Sorscher2022_BeyondScalingLaws, Feldman2020_CoreSetsSurvey} and hypothesised that unique points (which may be considered as edge cases) can have a significant impact on model performance  \citep{Feldman2020_LongTail}. One way to find such samples is by manually defining edge cases, e.g. under the involvement of domain experts, and actively collecting these data points. This can be extended by a combination of empirical and technical analysis, for instance, by plotting a downsampled embedding and analyzing the edge cases. A useful tool for this purpose is  \href{https://tensorleap.ai/}{tensorleap}. 
\pagebreak
Unfortunately, it is still unclear whether rare points are desirable or not in a dataset. While they can have a significant impact on model performance, they might make a model prone to privacy attacks \citep{Sorscher2022_BeyondScalingLaws, Feldman2020_LongTail}.

\subsubsection{Train-Test Split Creation}
Usually, after data collection the data would be labled and preprocessed. Since our suggestion regrading labeling are short, they are covered in the section on documentation. Apart from this, we have swapped the section on train-test split creation and preprocessing because it is recommend to do preprocessing on the splits separately whenever feasible.

\paragraph{Stratification}
\label{app:stratification}
First of all, training and test sets should be stratified.
If metadata was collected it should be considered for stratification. As indicated in Section \ref{subsec:data_collection}, another option is to create splits of different difficulty according to the metadata. An easy test set might contain overlapping instances between entities in training and test set (e.g., data of the same patients in training and test set) whereas a hard test set contains only data from  new entities (patients). This can be used to estimate the generalization capabilities of a model and helps to understand whether it is usable for a task or not. 

\paragraph{Test Set Size}
One of the most difficult choices is the size of the test set.
Despite the standard 80-20 (or similar) suggestion we are not aware of common strategies to estimate the necessary test set size in a general setting \citep{Joseph2022_OptimalRatioSplitting}. In medical and psychological research, sample size estimation based on statistical significance is established. For example, if the effect of medication against a certain disease should be evaluated and it is a priori known that the disease occurs in 10 from 1000 patients ($p=1\,\%$) and the target is an error margin of $\epsilon=$1\,\% with a confidence level of $\sigma=$ 99\,\%, the necessary sample size $n$ is given by 
$n \approx \frac{z_{0.99}^2\times p(1-p)}{\epsilon^2}=\frac{2.58^2 \times 0.01(1-0.01)}{0.01^2}=659$, 
where $z_{0.99}=2.58$ is the value for the selected confidence interval derived from the normal distribution. If $p$ is unknown, it is usually set to 0.5.
The same can be applied to \ac{ML} and was actually discussed in a few works especially from the medical domain \citep{Beleites2013_SampleSizePlanning, Dietterich_ApproximateStatisticalTestsSupervisedClassification, Raudys1990_SmallSampleSizeReccomendationPractitioners, Konietschke2021_SmallSampleSize}. 
In the case of ML, the interpretation is as follows; the goal is to evaluate an ML model that predicts the disease with a randomly drawn test set of size 659. Then, there is a 99\,\% confidence that the real world model performance is $\pm$1\,\% of the error rate on the test set. However, one drawback is that it is assumed that the collected data is representative of the real world setting and it is not guaranteed that the model will not learn any spurious correlations. 

Another indicator for a good test set is the test variance. The underlying assumption is that a better test set should show a low variance between different evaluation runs with the same model \citep{Bouthillier2021_Variance}. That is, if the same experiment results in an unexpectedly large variance, the test set is most likely too small and should be extended. An example for this effect is shown in Figure \ref{fig:performance_train_fraction}(b). Although multiple evaluations should generally be done, it is of special importance if the test set cannot be extended \citep{Bouthillier2021_Variance, Raschka2018_ModelEvaluationAndSelection, Lones2021_AvoidMLPitfalls}.

\subsubsection{Data Processing}
\label{subsec:data_processing}
After the data was collected and (maybe) labeled, it is preprocessed. This, for example, includes data cleaning, feature selection, and feature engineering. As mentioned above, it is important to do this separately on all splits whenever possible \citep{Lones2021_AvoidMLPitfalls, poretschkin2021_IAIS_Leitfaden}. Otherwise, information can leak into the test set. For example, by oversampling rare classes before creating splits \citep{winter2021_TueV_Austria}.  
In this work, we distinguish between processing steps that are a fixed part of a deployed dataset and as such is documented in the data card,
and steps specific to a particular \ac{ML} model which are documented in the model card. Generally, it is advisable to involve domain experts and to document whether and how the processing steps are agreed on with them \citep{poretschkin2021_IAIS_Leitfaden}.

\subsubsection{Understanding and Documenting Data \& Label Quality}
\label{subsec:data_quality}
Finally, the characteristics of the data need to be documented.
This includes all measures applied in order to understand the data. For example, plots of embeddings, identified rare points, or possible label errors or ambiguities. Although not focused on AI, the 16 dimensions defined in the \cite{ISO25012} standard are well suited for this. They include accuracy, completeness, consistency, credibility, currentness, accessibility, compliance, confidentiality, efficiency, precision, tractability, understandability, availability, portability, recoverability, and relevance. Dimensions that are of particular importance are accuracy (can be interpreted as fraction of label errors), completeness (fraction of missing values) and consistency (fraction of inconsistencies such as synonyms).
 
From the technical side, \href{https://cleanlab.ai/}{Cleanlab} \citep{Northcutt2021_LabelErrors} and \href{https://tensorleap.ai/}{tensorleap} are toolboxes that provide various methods for analyzing data and detecting and fixing label errors. Although it is not always possible to estimate whether something is a label error, documenting such ambiguous samples is important. Apart from this, the four-eyes principle is advisable when data is annotated manually \citep{poretschkin2021_IAIS_Leitfaden}. 
When dealing with third-party annotators, the usage of a \emph{gold standard dataset} is possible.
A gold-standard dataset is a small fraction of data with very high label quality that is send to an external provider for labeling and can be used to estimate the label quality of the provider afterwards. 

\newpage
{\small

\renewcommand\arraystretch{1.3}
\begin{longtable}{p{0.15\linewidth}|p{0.35\linewidth}p{0.15\linewidth}p{0.35\linewidth}}  
    \toprule
    \multicolumn{4}{c}{\bf{Data Card}}\\
    \toprule
    \bf{Step}        & \bf{Requirement}                                         &\bf{AI Act}       & \bf{References} \\
    \hline
    \hline
    \multirow{5}{=}{General}
                        & Name originator of the dataset and provide a contact person          & & \cite{Pushkarna2022_DataCards, Gebru2021_DataSheets}\\\cline{2-4}
                        & Describe the intended use of the dataset                             & & \cite{Pushkarna2022_DataCards, Gebru2021_DataSheets}\\\cline{2-4}
                        & Describe licensing and terms of usage                                & & \cite{Pushkarna2022_DataCards, Gebru2021_DataSheets} \\\cline{2-4}
                        
    \specialrule{.075em}{0em}{0em}
    \multirow{16}{=}{Data-description}
                        & Collect requirements for the data before starting data collection  & Art. 10 (2) d) e) \\\cline{2-4}
                        & Describe a data point with its interpretation                      & & \cite{Pushkarna2022_DataCards, Gebru2021_DataSheets} \\\cline{2-4}
                        & Maybe, provide additional documentation to understand the data (e.g., links to scientific sources, preprocessing steps or other necessary information) & & \cite{Pushkarna2022_DataCards}   \\\cline{2-4}
                        & If there is GDPR relevant data (i.e. personally identifiable information), describe it  & & \cite{Pushkarna2022_DataCards, Gebru2021_DataSheets}\\\cline{2-4}
                        & If there is biometric data, describe it                       &                   & \cite{Pushkarna2022_DataCards, Gebru2021_DataSheets, poretschkin2021_IAIS_Leitfaden}\\\cline{2-4}
                        & If there is copyrighted data, summarize it                    & Art. 28 b)        & \\\cline{2-4}
                        & If there is business relevant information, describe it        &                   & \cite{poretschkin2021_IAIS_Leitfaden}\\\cline{2-4}

    \specialrule{.075em}{0em}{0em}
    \multirow{16}{=}{Collection}
                        & Describe the data collection procedure and the data sources   & Art. 10 (2) b)  Annex III c)  & \cite{Pushkarna2022_DataCards, Gebru2021_DataSheets}\\\cline{2-4}
                        & Use data version control                                      & Art. 10 (2) b)    & \cite{poretschkin2021_IAIS_Leitfaden} \\\cline{2-4}
                        & Consider the prior requirements for the data                  & Art. 10 (2) e)    & \\\cline{2-4}
                        & Include and describe metadata                                 &                   & \href{https://www.dcc.ac.uk/guidance/standards/metadata}{Metadata Standards}\\\cline{2-4}
                        & Involve domain experts and describe their involvement         &                   & \cite{poretschkin2021_IAIS_Leitfaden} \\\cline{2-4}
                        & Describe technical measures to ensure completeness of data& Art.  10 (2) g) Art. 28 b) & \href{https://github.com/cleanlab/cleanlab/}{CleanLab}\\\cline{2-4}
                        & Describe and record edge cases                                &                   &  \cite{poretschkin2021_IAIS_Leitfaden, Pushkarna2022_DataCards} \\\cline{2-4}
                        & If personal data is used, make sure and document that all individuals know they are part of the data         & & \cite{Lu2021_DataDeprecation, Gebru2021_DataSheets} \\\cline{2-4}
                        & Describe if and how the data could be misused                 &                   & \cite{Gebru2021_DataSheets, Pushkarna2022_DataCards}\\\cline{2-4}
                        \pagebreak
                        \cline{2-4}
                        & If fairness is identified as a risk, list sensitive attributes & Art. 10 (5)  & Section \ref{app:protected_attrs}, \cite{ISO12791_Fairness} \\\cline{2-4}
                        & If applicable, address data poisoning                          &              & \cite{poretschkin2021_IAIS_Leitfaden}\\

    \specialrule{.075em}{0em}{0em}
    \multirow{4}{=}{Labeling}
                        & If applicable, describe the labeling process                   & & \cite{Gebru2021_DataSheets, Pushkarna2022_DataCards}, \href{https://snorkel.ai/}{snorkel.ai}\\\cline{2-4}
                        & If applicable, describe how the label quality is checked       & & Section \ref{subsec:data_quality}\\
                        
    \specialrule{.075em}{0em}{0em}
    \multirow{13}{=}{Splitting}
                        & Create and document meaningful splits with stratification      & Art.  10 (3)      &  Section \ref{app:stratification},  \cite{Beleites2013_SampleSizePlanning, Konietschke2021_SmallSampleSize, Raudys1990_SmallSampleSizeReccomendationPractitioners, Joseph2022_OptimalRatioSplitting, Zhang2022_SliceTeller, poretschkin2021_IAIS_Leitfaden} \\\cline{2-4}
                        & Describe how data leakage is prevented                                  &                   & \cite{poretschkin2021_IAIS_Leitfaden, winter2021_TueV_Austria}\\\cline{2-4}
                        & Recommendation: test the splits and variance via cross-validation       &                   & \cite{Bouthillier2021_Variance}\\\cline{2-4}
                        & Recommendation: split dataset into difficult, trivial and moderate      &                   & \cite{Meding2021_DDD} \\\cline{2-4}
                        & Recommendation: put special focus on label quality of test data         &                   & \cite{Northcutt2021_LabelErrors}\\\cline{2-4}
                        & Reminder: Perform separate data preprocessing on the splits             &                   & \cite{Roberts2017_CrossValidationTemporalSpatial, Lones2021_AvoidMLPitfalls}\\
                        
    \specialrule{.075em}{0em}{0em}
    \multirow{8}{=}{Preprocessing}
                        & Document and motivate all processing steps that are a fixed part of the data      & Art. 10 (2) c)    & Section \ref{subsec:data_processing} \\\cline{2-4}
                        & Document whether the raw data can be accessed                                     &                   & \cite{Gebru2021_DataSheets}\\\cline{2-4}
                        & If sensitive data is available highlight (pseudo)-anonymization                   & Art.  10 (5)      & \cite{poretschkin2021_IAIS_Leitfaden}\\\cline{2-4}
                        & If fairness is a risk, highlight fairness specific preprocessing                  & Art.  10 (5)      & \cite{poretschkin2021_IAIS_Leitfaden}\\
                        
    \specialrule{.075em}{0em}{0em}
    \multirow{9}{=}{Analyzing}
                        & Understand and document characteristics of test and training data.                                            &   Art. 10 (2) e) g), Art 10 (3) & Section \ref{subsec:data_quality}, \cite{ISO25012}, \cite{ISO5259}   \cite{Zhang2022_SliceTeller, Mitchell2022_MeasuringData, Mazumder2022_DataPerf} \\\cline{2-4} 
                        & Document why the data distribution fits the real conditions or why this is not necessary for the use case     & Art. 10 (4) & \cite{poretschkin2021_IAIS_Leitfaden}\\\cline{2-4}
                        & Document limitations such as errors, noise, bias or known confounders   & Art. 10 (2) f) g) & \cite{Gebru2021_DataSheets, Pushkarna2022_DataCards, Wittenbrink2022_LeitfadenQualitatsmanagementKI, poretschkin2021_IAIS_Leitfaden} \\

    \specialrule{.075em}{0em}{0em}
    \pagebreak
    \specialrule{.075em}{0em}{0em}
    \multirow{10}{=}{Serving}
                        & Describe how the dataset will be maintained in future                &               & \cite{Gebru2021_DataSheets, Pushkarna2022_DataCards}, \cite{Lu2021_DataDeprecation}\\\cline{2-4}
                        & Describe the storage concept (e.g., everything users need to know to access the data). For developers it must be possible to document on which version of the data a specific model was trained.                                 &               & \cite{poretschkin2021_IAIS_Leitfaden}\\\cline{2-4}
                        & Describe the backup procedure                                        &               & \cite{poretschkin2021_IAIS_Leitfaden}\\\cline{2-4}
                        & If necessary, document measures against data poisoning               &               & \cite{poretschkin2021_IAIS_Leitfaden}\\

    \specialrule{.075em}{0em}{0em}
    \multirow{2}{=}{Further notes}       
                        & Document further recommendations or shortcomings in the data \\\cline{2-4}
                        
    \specialrule{.075em}{0em}{0em}
    \multirow{2}{=}{Involvement of Individuals}                            
                        & If feasible, describe involvement of affected individuals            & Art. 29 5a. \\     

    \bottomrule
  \end{longtable}
}

\pagebreak 

\subsection{Model Development}
Developing models that execute a specific task in a reliable and fair way is challenging. The AI Act phrases the requirements for (high-risk) AI systems in Article 15 (1): \emph{``High-risk AI systems shall be designed and developed in such a way that they achieve [...] an appropriate level of accuracy, robustness and cybersecurity, and perform consistently in those respects throughout their lifecycle.''} To achieve this, there are several concerns such as explainability, fairness, feature engineering and others. 
For documenting all these pillars, we build on the work of \citet{Mitchell2019_ModelCards}. While some points overlap, we added  and complemented others, i.e., \ac{XAI} and testing in real world settings. 

\subsubsection{Explainability}
First, the requirements for \ac{XAI} should be determined. In this regard, different aspects of explainability methods needs to be considered.
The first important aspect is to distinguish between the need to explain a single decision (local explainability) or the entire ML model (global explainability).
The first might be important in cases where lay users need to be able to understand a decision made about them (e.g., financial advise), while the latter might be more important to make sure a model does not decide based on sensitive features.
These explanations can be derived via different means, with most taxonomies listing post-hoc explainability vs. explainability by nature or design, also so-called \ac{IML}.
The former describes using a black-box model for a decision and generating explanations afterwards, most often via simpler approximations, with the most prominent examples being LIME \citep{ribeiro2016_lime} and SHAP \citep{lundberg2017_shap}.
For post-hoc explainability, it is important to note that most \ac{XAI}-methods provide no formal guarantees, thus explanations might not conform to the ML-model's decision in all cases.
Another important factor and currently open problem is how to evaluate such explanation methods. While many evaluation metrics and tests have been proposed, none has been widely adopted and consistently been used throughout the literature \citep{nauta2022_XAIreview, Burkart_Explainability}.
Currently it seems that explanation methods should be tested in real-world scenarios, including people with the same explainability needs as the target group of the explanations, as e.g. ML-engineers might need different explanations than lay users.
Additionally to the requirements regarding explainability and transparency itself, \ac{XAI}-methods might be needed to fulfill other legal requirements.
If, for example, users of an ML-system have a right to object to decisions (as given through the \ac{GDPR} in Europe), they must be sufficiently informed of a decision to be able to make use of their rights.
In general, we advise to think of explainability demands before choosing a model, as use cases with a high demand for transparency might need to provide some guarantees for correct explanations and could thus be limited to \ac{IML} models.
Furthermore,  considerations regarding explainability can also influence other development steps, especially data preprocessing.
Here, non-interpretable features as for example obtained by \ac{PCA} can severely limit the expressiveness of explainability approaches.
Regarding the use of explainability for AI systems, norms are currently specified, but not yet published \citep{DIN-Spec-92001-3,ISO6254-explainability}.

\subsubsection{Feature Engineering, Feature Selection, and Preprocessing}
While some of the preprocessing is addressed in the data card, here we consider only the processing that is not part of the dataset itself, but is model specific. Feature engineering and selection as well as data augmentation and preprocessing steps should always be meaningful in the context of the use case and the used model. Feature selection can be seen as an optimization problem to find the most relevant features for the model. While the performance of deep learning algorithms might not change, the computation cost will increase and consume time and resources \citep{cai2018_featureSelection}. To prevent data leakage, all steps, including feature selection, feature engineering, and preprocessing, should be performed after splitting the data \citep{Kapoor22_LeakageAndReproducibility, reunanen2003variableselection}. This also holds for k-fold cross validation, where feature engineering and preprocessing must be executed on each fold independently.

\subsubsection{Model Selection}
The range of usable models depends on the given use case and its requirements such as on explainability, prediction performance and inference time, just to name a few. With those limitations, the amount of meaningful models will narrow down. It is good practice to compare the selected model to standard baselines. A good starting point are easy to train  and for the use case established state of the art models. Also, rule-based systems or heuristic algorithms can be used as a baseline. Trying different classes of models that have shown good results on a given set of problems is always recommended \citep{Lones2021_AvoidMLPitfalls, ding2018model}. According to the no free lunch theorem, there is no single machine learning algorithm that can outperform all others across all possible problems and datasets \citep{wolpert2002supervised}.
The choice of the algorithm needs to be appropriate for the underlying problem class and the amount and quality of the available data. Usually, there are fine-tuned algorithms for  natural language processing, computer vision, time series and other fields. Note that deep learning models will not always be the best option. \cite{grinsztajn2022tree} show that tree based models often outperform deep learning in tabular data. In time series forecasting, there is no clear best algorithm class. Statistical and tree based methods regularly are on par with deep learning models  \citep{zeng2022transformers}. However, achieving this improved performance often requires a significant increase in computational resources \citep{makridakis2023statistical}. In general, simpler models are preferred over more complex models if they provide similar levels of performance, according to the principle of Ockham's razor, because they are less prone to overfitting and easier to interpret.

\subsubsection{Metrics}
Similar to model choice, the metrics should also be selected carefully. Depending on the use case,  recall or precision might be more appropriate metrics than general ones like accuracy or F1-score \citep{poretschkin2021_IAIS_Leitfaden}. The choice of model also influences the choice of useful metrics  \citep{naser2021_metrics, ferri2009metrics}. Furthermore, the minimum acceptable score for the selected metrics should be determined to create value in the specific use case. It is always recommended to involve domain experts to address the minimum requirements.

\subsubsection{Hyper-parameter Optimization}
\ac{HPO} is usually the most computational heavy task in model development with up to hundreds of hyper-parameters to tune. It can help to understand the underlying problem and the used algorithms to be able to reduce the space for the optimal set of hyper-parameters. For large sets of hyper-parameters there are advanced optimization techniques like Bayesian optimization and early stopping. \citet{feurer2019hyperparameter} and \citet{yang2020_hyperparameter} give good overviews on this.

Furthermore, \ac{HPO} is another pitfall of leakage. It is best to ``lock'' the test set to prevent leakage, i.e., to not use the test data until \ac{HPO} is completed. In most cases, using (stratified) k-fold cross-validation is the best practice. However, cross-validation and methods for \ac{HPO} might introduce some unwanted variance \citep{Bouthillier2021_Variance, Wong2019_ReliableCV, yang2020_hyperparameter, bischl2021_hyperparameter}. 

\subsubsection{Model Evaluation}
To evaluate model performance, prevent over-fitting or under-fitting, and test generalizability, models must be evaluated against test data. As stated above, the test set must not be used for \ac{HPO} or any kind of ``intermediate evaluation''. The performance should be evaluated against the previously defined minimum requirements for the use case. If the performance of the model is low, there are many possible causes, e.g., noisy data, model choice, or metrics. If the model over-performs on the test set, i.e., the test error is smaller than the training error, one should be cautious of potential errors such as leakage. \citet{zhang2020_MLtesting, Braiek2018_MLtesting} give an overview over the topic. On the standardization side, \cite{ISO4213} addresses the classification performance of \ac{ML} models.

\subsubsection{Model Confidence}
\citet{Guo2017_Calibration} show that the confidence values calculated by modern \ac{ML} models is often poorly calibrated, i.e., the underlying correct likelihood of a  class is not represented by the predicted probability of the model. Even though this might not be a big issue in many cases, it can lead to unwanted behavior of models in production and unstable results. Therefore, confidence values should be evaluated and calibrated if necessary \citep{Guo2017_Calibration, abdar2021_uncertainty}. 
Instead of calibrating models post-hoc, conformal prediction allows to train algorithms with a fixed confidence \citep{shafer2008tutorial}.

\subsubsection{Testing in Real-World Setting}
Apart from model evaluation on the test set, a model must be tested and benchmarked in the real-world setting \citep{Arp_DosDonts, larson2021_regulatory}. Domain experts should be involved in the testing design, and it should ideally be done before the first model is trained. 
The design should consider edge cases, disturbances, changes of the environment, and all other likely settings and changes. Testing in a real-world setting will also uncover possible leakage and other shortcomings that might slip the model evaluation. 
If real world testing is not possible, computational stress tests should be done \citep{young2021_stresstest}.

{\small
\renewcommand\arraystretch{1.3}
\begin{longtable}{p{0.15\linewidth}|p{0.35\linewidth}p{0.15\linewidth}p{0.35\linewidth}} 
    \toprule
    \multicolumn{4}{c}{\bf{Model Card}}\\
    \toprule
    \bf{Step}   & \bf{Requirement}   &  \bf{AI Act}   & \bf{References} \\
    \hline
    \hline
    \multirow{6}{=}{General}    
            & Provide Name and details of a contact person            \\\cline{2-4}
            & Provide Name of the person who created the model      &   & \cite{Mitchell2019_ModelCards}\\\cline{2-4}
            & Document the creation date and version of the model           &   & \cite{Mitchell2019_ModelCards}\\\cline{2-4}
            & Describe intended use of the model              &   & \cite{Mitchell2019_ModelCards}\\
            
    \specialrule{.075em}{0em}{0em}
    \multirow{10}{=}{Description}       
            & Describe the architecture of the used model    &   & \cite{Mitchell2019_ModelCards}\\\cline{2-4}
            & Describe the used hyper-parameters           &   & \\\cline{2-4}
            & Document the training, validation, and test error     &   & \cite{Mitchell2019_ModelCards}\\\cline{2-4}
             & Document computation complexity, training time and energy consumption (for foundation models describe steps taken to reduce energy consumption) & Art. 28 & \cite{yang2017energy, garcia2019energy}\\
                        
    \specialrule{.075em}{0em}{0em}
    \multirow{6}{=}{Explainability and interpretability}    
            & Document the demand for explainability and interpretability & Art. 13 (1, 2), Art. 15 (2)\\\cline{2-4}
            & Describe taken actions if any & Art. 15 (2) & \cite{ISO6254-explainability, ribeiro2016_lime, nauta2022_XAIreview, Schaaf_XAI, Burkart_Explainability}\\     

    \specialrule{.075em}{0em}{0em}
    \multirow{7}{=}{Feature engineering, feature selection and preprocessing}   
            & Describe the consideration of explainability and interpretability (if relevant) & Art. 10 (3, 4) & \\\cline{2-4}
            & Describe feature engineering and selection with (domain specific) reasoning   &   & \cite{Mitchell2019_ModelCards, Kapoor22_LeakageAndReproducibility, cai2018_featureSelection}\\\cline{2-4}
            & Describe and reason of preprocessing steps  &   & \cite{Mitchell2019_ModelCards, Kapoor22_LeakageAndReproducibility}\\

    \specialrule{.075em}{0em}{0em}
    \pagebreak
    \specialrule{.075em}{0em}{0em}
    \multirow{18}{=}{Model selection}   
            & Describe the consideration of explanibility and interpretability (if relevant)\\\cline{2-4}
            & Describe the base line model and its evaluation   &       & \cite{Lones2021_AvoidMLPitfalls}\\\cline{2-4}
            & Describe the reasoning for the model choice       &       & \cite{Lones2021_AvoidMLPitfalls} \\\cline{2-4}
            & Describe why the complexity of model is justified and needed & & \\\cline{2-4}
            & Document the comparison to other considered models \\\cline{2-4}
            & Document the approach of hyper-parameter optimization   &   &   \cite{Bouthillier2021_Variance, Wong2019_ReliableCV, yang2020_hyperparameter, feurer2019hyperparameter, tuningplaybookgithub, Raschka2018_ModelEvaluationAndSelection}\\\cline{2-4}
            & Describe the model evaluation  & Art. 5, 6, 7, 9    & \cite{Zhang2022_SliceTeller, zhang2020_MLtesting, Braiek2018_MLtesting, Raschka2018_ModelEvaluationAndSelection, ISO4213}\\
            
    \specialrule{.075em}{0em}{0em}
    \multirow{6}{=}{Choice of metrics}  
            & Describe selected metrics and describe reasoning regarding use case and fairness    &  & \cite{Mitchell2019_ModelCards, Wachter2021_BiasPreservation, ferri2009metrics, poretschkin2021_IAIS_Leitfaden, ISO12791_Fairness}\\\cline{2-4}
            & Define the minimum requirement for use in production (domain specific reasons)\\
            
    \specialrule{.075em}{0em}{0em}
    \multirow{3.5}{=}{Model confidence}     
            & Document and quantify uncertainty of the model    &   & \cite{abdar2021_uncertainty}\\\cline{2-4}
            & Document approach of dealing with uncertainty     &   & \cite{Guo2017_Calibration, shafer2008tutorial}\\

    \specialrule{.075em}{0em}{0em}
    \multirow{7}{=}{Testing in real world setting}      
            & Describe the test design &  Art. 5, 6, 7, 9  & \\\cline{2-4}
            & Describe possible risks, edge cases and worst case scenarios and create (or simulate) them if possible & Art. 15 (3), Art. 28\\\cline{2-4}
            & Describe limitations and shortcomings of the model & & \\\cline{2-4}
            & Describe the test results\\\cline{2-4}
            & Explain the derived actions\\

    \specialrule{.075em}{0em}{0em}
    \multirow{2}{=}{More}       
            & Describe further recommendations or shortcomings of the model \\
            
    \specialrule{.075em}{0em}{0em}
    \multirow{2}{=}{Involvement of Individuals}     
            & If feasible, describe involvement of affected individuals & Art. 29 (5) a) & \\     
    \bottomrule
\end{longtable}
}
\pagebreak
\subsection{Model Operation}
\label{subsec:monitoring}
Proper handling of AI applications during deployment involves managing different aspects. Technical correctness and desired performance of the AI application is one, but IT security, maintenance and MLOps, workforce acceptance, knowledge management, and legal concerns must also be considered. Lawmakers are very clear in their demands \citep{AiAct, EU2019_Ethic}. Article 61 (1) of the AI Act states \emph{``Providers shall establish and document a post-market monitoring system in a manner
that is proportionate to the nature of the artificial intelligence technologies and the
risks of the high-risk AI system.''} The challenge is to establish specific methods to control assessed risks. Ultimately, decisions need to be made individually for each use case. We give an overview over currently discussed topics and introduce our monitoring card as a de facto checklist. 

\subsubsection{Operating Concept}
In order to use an AI application in productive operation, the operational concept defines the way the application interacts with existing processes. The AI Assessment Catalog specifies this to a detailed description of the used software components and APIs to other environments \citep{poretschkin2021_IAIS_Leitfaden}. As AI falls under the category of software, \cite{winter2021_TueV_Austria} suggest similar IT security precautions, as required for other high-impact software. Additionally, the \cite{ISO27000-Cyber-Security} series concerning Cyber Security and the more recent \cite{ISO27001-ISMS} provide valuable recommendations and best practice examples for establishing an IT security management system (ISMS). Furthermore, it is beneficial to draw upon recommendations from classical software engineering practices when formulating the operating concept. \cite{SCHMIDT2013swengineering} state that \emph{``Concepts of operation descriptions usually address the following:}
\emph{
\begin{itemize}
    \item[1.] Statement of the goals and objectives of the software product.
    \item[2.] Mission statement that expresses the set of services the product provides.
    \item[3.] Strategies, policies, and constraints affecting the product.
    \item[4.] Organizations, activities, and interactions among participants and stakeholders.
    \item[5.] Clear statement of responsibilities associated with product development and sustainment.
    \item[6.] Process identification for distributing, training, and sustainment of the product.
    \item[7.] Milestone decision definitions and authorities.''
    \end{itemize}
    }
\noindent Extending these requirements, the AI Act mandates in Art. 13 the disclosure of the use of AI in outcome determination to users, with the level of detail depending on their function. In all cases, it is essential to explain the reasoning behind AI's decisions and the impact of incoming data features, while also implementing a process to override decisions when necessary. Additionally, for both regular and emergency tasks, a failsafe responsibility must be established. Finally, it is recommended to provide training to employees on the usage of the AI application to enhance acceptance and productivity.

\subsubsection{Monitoring}
To ensure technical correctness, continuous evaluation of the AI application is essential. Depending on the risk level, it is advisable to establish either live monitoring or periodic controls. Live monitoring of the application is recommended to quickly detect anomalies, not only for emergency situations but also to enhance AI performance. The AI Act, as specified in Article 9, mandates the implementation of a `risk management system' that assesses risks during deployment and evaluates ongoing mitigation strategies through an iterative process. The objective is to identify conditions under which the AI system can operate safely. Deviating from this realm of safe operation necessitates the deactivation of the AI application and a transition to a failsafe software system.

\subsubsection{Human-Computer-Interface and Transparency}
For monitoring purposes and insights into the functioning of the AI, a suitable interface is required. AI Act Art. 14 specifies: \emph{``It has to be guaranteed that the potentials, limitations and decisions are transparent without bias.''} It is proposed that the interface provides information at different levels of detail. This way operating staff can be supported in carrying out their tasks with crucial process relevant information by the AI, while data scientists might want to get more insights into training meta data to optimize learning curves. Avoiding bias is a current research topic. \citet{kraus2021erklarbare} point out that an explanation of the algorithm alone is not sufficient for transparency. In accordance with the recommendations set forth by \cite{Unesco2022}  \emph{``Transparency   aims   at   providing   appropriate   information to the respective addressees to enable their understanding and foster trust. Specific to the AI system, transparency  can  enable  people  to  understand  how  each  stage  of  an  AI  system  is  put  in  place,  appropriate  to  the  context  and  sensitivity  of  the  AI  system.  It  may  also  include  insight  into  factors  that  affect  a  specific  prediction  or  decision,  and  whether  or  not  appropriate  assurances  (such  as  safety  or  fairness  measures)  are  in  place. In cases of serious threats of adverse human rights impacts,  transparency  may  also  require  the  sharing  of  code or datasets.''}

\subsubsection{Model Drift}
Model drift summarizes the impact of changes in concept, data, or software and their effect on the model predictions. Concept drift can occur when the underlying correlations of input and output variables change over time, while data drift can occur when the known range is left, and software drift can occur when, for example, the data pipeline is modified and data fields are renamed or units are changed. The effects of changes in the model ecosystem should be closely monitored. That is, to prevent a low model performance based on corrupted or unexpected input data, it is best practice to compare the input data with training data. Metrics such as f-divergence may be used to express differences in the data distribution and calculate limits for how far the live model performance may deviate from the training performance.

\subsubsection{Privacy}
If personal data is collected in the course of the AI deployment, secure storage of this data and deletion periods must be established with the help of a data protection officer. In addition, a process for providing information about personal data must be created in order to process requests from data subjects. The \ac{GDPR} imposes a minimization of personal data collection. Methods such as differential privacy may be used to privatize personal data during collection or procession  \citep{dwork2014algorithmic}.

\subsubsection{MLOps}
While the AI Act imposes no specific requirements for MLOps, except for ensuring the safe operation of AI throughout its lifecycle,  \cite{poretschkin2021_IAIS_Leitfaden} provide several recommendations regarding the implementation of secure MLOps practices.

To ensure uninterrupted operation, a well-regulated MLOps process is of utmost importance. The frequency of regular updates can be determined based on risk and technology impact assessments. When there is a higher risk associated with deviations between the training data and model data, more frequent retraining should be considered. It is crucial to store metadata related to the training process and model, including parameters used, training data, and model performance, to enable reproducibility. This approach relies on the availability of rapidly collected and accurately labeled new data. In situations where such data collection and labeling are not feasible, additional oversight of the AI system by trained personnel, such as data scientists, becomes necessary.

Moreover, based on the identified risks, it is advisable to establish a dedicated testing period for updates to both the model and infrastructure. Smooth transitions necessitate clear delineation of responsibilities for the commit and review processes, as well as deployment. In cases where swift rollback is required, well-documented versioning including data is essential. Additionally, employing a mirrored infrastructure enables updates to be installed seamlessly without causing downtime.

\newpage

\small{
\renewcommand\arraystretch{1.3}
\begin{longtable}{p{0.15\linewidth}|p{0.35\linewidth}p{0.15\linewidth}p{0.35\linewidth}} 
    \toprule
    \multicolumn{4}{c}{\bf{Operation Card}}\\
    \toprule
    \bf{Step}           & \bf{Requirement}   &  \bf{AI Act}   & \bf{References} \\
    \hline
    \hline
    \multirow{4}{=}{Scope and aim of monitoring}
                                    & Describe monitored components & Art 61 (1,3)\\\cline{2-4}
                                    & Assess risks and potential dangers according to Use Case Card & Art 9 (2) \\\cline{2-4}
                                    & List safety measures for risks & Art. 9 (4) &\cite{poretschkin2021_IAIS_Leitfaden} \\
                                    
    \specialrule{.075em}{0em}{0em}
    \multirow{4}{=}{Operating concept}
                                    & Create and document utilisation concept & Art. 13, Art. 17, Art. 16 \\\cline{2-4}
                                    & Plan staff training & Art. 9 (4c) & \cite{BMWK2017industrie} \\\cline{2-4}
                                    & Determine responsibilities & &\cite{poretschkin2021_IAIS_Leitfaden}\\
    \specialrule{.075em}{0em}{0em}
    \multirow{4}{=}{Autonomy of application}
                                    & Document decision-making power of AI & Art. 14, Art. 17 & \cite{poretschkin2021_IAIS_Leitfaden, EU2019_Ethic} \\\cline{2-4}
                                    & Determine process to overrule decisions of the AI & Art. 14 (4) d) e) & \cite{poretschkin2021_IAIS_Leitfaden}\\
    
    \specialrule{.075em}{0em}{0em}
    \multirow{3}{=}{Responsibilities and measures} 
                                    & Document component wise: Assessed risk, control interval,              responsibility, measures for emergency & Art. 9 & \cite{poretschkin2021_IAIS_Leitfaden, EU2019_Ethic, BMWK2017industrie} \\

    \specialrule{.075em}{0em}{0em}
    \multirow{5}{=}{Model performance}
                                    & Monitor input and output & Art. 17 (1) d), Art. 61(2) & \cite{poretschkin2021_IAIS_Leitfaden}\\\cline{2-4}
                                    & Detect drifts in input data & &\cite{poretschkin2021_IAIS_Leitfaden}\\\cline{2-4}
                                    & Document metric in use to monitor model performance & Art. 15 (2) \\
                                    
    \specialrule{.075em}{0em}{0em}
    \multirow{4}{=}{AI interface}
                                    & Establish transparent decision making process & Art. 52 & \cite{poretschkin2021_IAIS_Leitfaden, kraus2021erklarbare} \\\cline{2-4}
                                    & Establish insight in model performance on different levels & Art. 14 & \cite{poretschkin2021_IAIS_Leitfaden} \\
    
    \specialrule{.075em}{0em}{0em}
    \multirow{6}{=}{IT security}
                                    & Document individual access to server rooms & Art. 15 & \cite{winter2021_TueV_Austria, BMWK2019AIandLaw}\\\cline{2-4}
                                    & Set and document needed clearance level for changes to AI/deployment/access regulation & &\cite{winter2021_TueV_Austria, BMWK2019AIandLaw} \\\cline{2-4}
                                    & Establish and audit ISMS & & \cite{ISO27000-Cyber-Security, ISO27001-ISMS}\\

    \specialrule{.075em}{0em}{0em}
    \multirow{4}{=}{Privacy}
                                    & Justify and document use or waiver of a privacy preserving algorithm & Art. 10 (5) & \cite{poretschkin2021_IAIS_Leitfaden, dwork2014algorithmic}\\\cline{2-4}
                                    & If applicable, document privacy algorithm and due changes in the monitoring of output data & &\cite{dwork2014algorithmic} \\ 
    \specialrule{.075em}{0em}{0em}
    \pagebreak
    \specialrule{.075em}{0em}{0em}
    \multirow{6}{=}{MLOps}          
                                    & Establish versioned code repository of AI, training and deployment & &\cite{poretschkin2021_IAIS_Leitfaden} \\\cline{2-4}
                                    & Establish maintenance and update schedule & &\cite{poretschkin2021_IAIS_Leitfaden} \\\cline{2-4}
                                    & Set regulation for the retraining of the AI and decision basis for the replacement of a model & & \cite{poretschkin2021_IAIS_Leitfaden} \\

    \specialrule{.075em}{0em}{0em}
    \multirow{6}{=}{Testing and rollout}
                                    & Determine responsibilities for updates & &\cite{poretschkin2021_IAIS_Leitfaden}\\\cline{2-4}
                                    & Document software tests & Art. 17 (1) d) & \cite{poretschkin2021_IAIS_Leitfaden} \\\cline{2-4}
                                    & Set period of time that an update must function stably before it is transferred to live status \\    
    \specialrule{.075em}{0em}{0em} 
    \multirow{2}{=}{Involvement of
Individuals}
                                    & If feasible, describe involvement of affected individuals & \\     

    \bottomrule
\end{longtable}
}

\pagebreak

\section{Summary and Conclusion}
We introduce a guideline for documenting AI applications along the entire developing process that extends previous work on model and data cards. For this purpose, we held interviews with certification experts, developers of AI applications preparing for the upcoming regulation of the European Union, and, individuals affected by the introduction of an AI system. All their feedback was incorporated into our framework. 
In contrast to works on model and data cards, we cover the entire development process, which results in defining two new cards: use case card and operation card. Further, we add a perspective on regulation. Our work is also different from most regulation and certification efforts by being more specific and following the development process of AI applications instead of risk dimensions.  Our guideline is meant to support the development of trustworthy and easy to audit AI systems from the beginning and should lay the foundations for a future certification of a system. Although AI regulation is still in an early phase and many changes may be expected, the main foundations and associated risks seem to have emerged. Hence, this work can be seen as a snapshot of the current state and will evolve over time but should already position practitioners well when legal requirements come into force.

While our guideline is more specific than many previous works, both the field of AI but also the regulation are progressing fast. Hence, certain changes or specifications can be expected. For example, best practices for creating datasets are likely to change in future. For this purpose, a structured and regularly updated collection of best practices and safety measures applied by different organizations similar to the \href{https://incidentdatabase.ai/}{AI incidents database} or the \href{https://appliedaiinitiative.notion.site/Risk-Classification-Database-2b58830bb7f54c9d8c869d37bdb27709}{Risk-Classification Database} could be collected.

\acks{This work builds on insight from the research projects \href{https://www.ipa.fraunhofer.de/en/reference_projects/veoPipe_Reliable_AI_product_development_process.html}{veoPipe} and \href{https://www.ipa.fraunhofer.de/en/reference_projects/ML4Safety.html}{ML4Safety} and a collaboration with Audi AG. We thank Hannah Kontos and Diana Fischer-Preßler for proofreading.}

\appendix
\clearpage

\section{Toy Audit}
\label{app:audit}
We exemplary applied our cards to one real world use case at our industry partner. However, in order to avoid revealing confidential information we present results of a toy audit here in the paper.
Consider a large bicycle manufacturer producing around 50\,000 pieces per year. The manufacturer plans to upgrade its product portfolio with carbon fiber frames due to their lightness. The number of bicycles with carbon frames is estimated to be around 5\,000 pieces per year. However, carbon frames have one drawback: they can break easily if something went wrong during the production process. For this reason, the manufacturer plans to use a two-fold quality inspection consisting of a regularly reoccurring computed tomography that can detect production errors reliably but at a very high cost every 100 pieces. On top of that, each frame shall undergo a visual quality inspection using computer vision before shipping (binary classification problem into defect and not-defect parts). If the computer vision (CV) tool predicts a defect, a worker will double check the frame.

\newpage
{\small
\begin{longtable}{p{0.15\linewidth}|p{0.35\linewidth}p{0.5\linewidth}}  
    \toprule
    \multicolumn{3}{c}{\bf{Use Case Card Part 1}}\\
    \toprule
    \bf{Step} & \bf{Requirement} & \bf{Measures} \\
    \hline
    \hline
    General 
        & Name contact person                 & John Doe, Project Manager \\\cline{2-3}  
        & List groups of people involved      & John Doe, Jane Roe, Richard Miles \\\cline{2-3}  
        & Summarize the use case shortly      & Produced carbon fiber bicycle frames shall undergo a quality inspection. Using computed tomography (CT) scans yields high precision, but would be associated with high costs. To reduce costs a machine learning model shall perform automated visual quality inspections on every frame and expensive CT-scans will be performed less. If a defect is predicted, the frame will be inspected by a trained professional. To ensure quality standards every \textit{100}-th frame will be inspected with a CT image. \\\cline{2-3}  
        & Describe the status quo             & To this day carbon fiber bicycle frames can only be inspected manually or by CT-scans. Current non carbon fiber frames are stress-tested and inspected during quality end control. During the initial year 5\,000 carbon fiber frames were produced and manually inspected, with 300 defective parts detected. For a larger database, an additional 200 images of defective frames were obtained from a service partner. \\\cline{2-3}  
        & Describe the planed interaction of sub-components (e.g., different software modules or hardware and software)   & After the pressing process fabricated parts will be photographed by industrial high speed cameras and evaluated by the machine learning model. The predictions of the ML model will be visualized with an interface for quality management workers, predicted defects will be highlighted and require a manual inspection result. For each frame produced, a randomized algorithm determines whether a CT-scan should be performed. Shift workers will then initiate the CT-scan process and document the result.   \\\cline{2-3}  
        & Provide a short solution summary    & The business problem can be translated to a binary classification problem, which can be solved by means of industrial CV. \\
        
    \hline
    Problem definition 
        & Clearly describe the learning problem  & Carbon fiber frames are highly vulnerable to tears and deformations, which can result in stability issues and thus, braking of the frame. In use this could cause accidents and harm to cyclists. Manual and CT-scan inspections are too costly to be performed for every frame. A solution that uses algorithms must work accurately, but fast enough not to delay production times. \\\cline{2-3}  
        \pagebreak
        \cline{2-3}  
        & Describe the disadvantages of the current approach & High costs for CT-scans, manual visual inspections are not feasible due to high time consumption and as a mundane task not appealing to workforce.\\\cline{2-3}  
        & Argue why classical approaches are not sufficient & Rule-based algorithms may not yield the necessary confidence. \\
        
    \hline
    Solution approach 
        & Describe the integration into the current workflow           & Images of the frames are automatically taken. The ML model will evaluate the images in a few seconds and present the results to current shift workers via an interface. Detected defective frames will automatically be marked for manual inspection. The pass/fail labels assigned by the model can  be manually overwritten by workforce. Workers will perform manual inspections on defective parts before submitting the frames to further assembly.\\\cline{2-3}  
        & Formulate the learning problem &  Binary label classification with optional heatmaps / attribution maps applied to the predictions.\\\cline{2-3} 
        & Formulate the KPIs for go-live & A recall above 99\% is targeted while preserving a precision of more than 90\% and the maximum inference time should not exceed 1 second.\\\cline{2-3}  
        & Provide a short data description & Photos are taken with an industrial camera. Both sides of the frame are photographed. The pictures are of size 1024 x 1024 pixels.\\
    \hline
\end{longtable}
}

\newpage
{\small
\begin{longtable}{p{0.15\linewidth}|p{0.35\linewidth}p{0.5\linewidth}}  
    \toprule
    \multicolumn{3}{c}{\bf{Use Case Card Part 2}}\\
    \toprule
    \multicolumn{3}{c}{Risk Assessment} \\
    \hline
    \hline
    Risk Class          
        & Categorize the application into a risk class according to the AI Act & Low risk. Although the carbon fiber frame is a safety relevant part of the bike, the AI is used for quality control and not part of the final product.\\
        
    \hline
    Human agency and oversight 
        & Rate the level of autonomy & Low risk since the AI only gives a suggested quality label. Manual overwrite is always possible. \\
        
    \hline
    Technical robustness and safety 
        & Evaluate the danger to life and health & Miss-classified broken frames might be shipped and customers may be hurt. \\\cline{2-3}  
        & Identify possibilities of non-compliance & No risk, standard safety requirements are fulfilled without the use of an AI, the AI is an additional layer of security. \\\cline{2-3}  
        & Identify customer relevant malfunction & No risk, the AI is solely used for internal quality control. \\\cline{2-3}  
        & Identify internal malfunction & Low risk of production stoppage or higher production costs. \\
        
    \hline
    Privacy and data governance     
        & List risks connected with customer data & No risk (no data collected). \\\cline{2-3}  
        & List risks connected with employee data & Risk of analyzing quality of work of production staff. \\\cline{2-3}  
        & List risks connected with company data & Low risk of production flaws being leaked to the outside world. \\
        
    \hline
    Transparency                    
        & Evaluate effects of incomprehensible decisions or the use of a black box model & Might result in trust issues in the AI within the workforce.                                 This will lead to either more manual inspections and with that higher costs or fewer inspections and thus lower quality.  \\\cline{2-3}  
        
    \hline
    Diversity, non-discrimination and fairness 
        & Check for possible manipulation of groups of people & No risk.  \\\cline{2-3}
        & Check for discrimination of groups of people regarding sensitive attributes  & No risk. \\
        
    \hline
    \pagebreak
    \hline
    Societal and environmental well being 
        & List possible dangers to the environment & No risk. Good quality assurance reduces waste.\\\cline{2-3}
        & Consider ethical aspects & No risk. \\\cline{2-3}  
        & Evaluate effects on corporate actions & An ELSI-Group (Ethical, Legal and Social Implications) has been formed by the workers council. \\\cline{2-3}  
        & Identify impact on the staff & The AI is solely used to ensure quality in the product. No devaluation of workforce quality is practiced. No analyses are conducted on the quality of work of the production staff and no employment contracts are terminated due to the use of AI. \\    
        
    \hline
    Accountability                      
        & Estimate the financial damage on failure & Broken frames might be shipped and need to be replaced. If customers get injured they might file for                             lawsuits. Misclassified flawless frames might get disposed of. \\\cline{2-3}  
        & Estimate the image damage on failure & Shipped misclassified broken frames might associate the company with poor quality. \\
        
    \hline
    Norms               
        & List of relevant norms in this context & \href{https://www.iso.org/ics/43.150/x/}{ISO 43.150 Cycles} \\
    \bottomrule
  \end{longtable}
}

\newpage
{\small
\begin{longtable}{p{0.15\linewidth}|p{0.35\linewidth}p{0.5\linewidth}}  
    \toprule
    \multicolumn{3}{c}{\bf{Data Card}}\\
    \toprule
        
    \bf{Step}            & \bf{Requirement}                                & \bf{Measures} \\
    \hline
    \hline
    
    General        
        & Name originator of the dataset and provide contact person            & John Doe, Project Manager.  \\\cline{2-3}               
        & Describe the intended use of the dataset                     & Train a supervised ML model for the recognition of defective bike frames.   \\\cline{2-3}              
        & Describe licensing and terms of usage                     & Part of the data was recorded internally and can be used without any restrictions. Another part of the data is provided from a service partner under a specific licensing agreement. \\\cline{2-3}             
    
    \hline
    Data-Description  
        & Collect requirements for the data before starting data collection & Before starting the data collection process, domain experts where involved to discuss how defects occur on the bikes. We discussed the proper amount of data and the necessary resolution of the images.\\\cline{2-3}
        & Describe a data point with its interpretation             & One data point consists of two RGB images with a  resolution of 1024$\times$1024 pixels. One image shows the frame frame from the left, the other one from the right. Each image was taken with an industrial camera of type [MODEL].  \\\cline{2-3} 
        & Maybe, provide additional documentation to understand the data & Not necessary, image data are human interpretable.        \\\cline{2-3}
        & If there is GDPR relevant data. describe it  & No personal data.               \\\cline{2-3}
        & If there is biometric data, describe it                   &   No biometric data.               \\\cline{2-3}
        & If there is copyrighted data, summarize it                &   No copyrighted data.               \\\cline{2-3}
        & If there is business relevant information, describe it          & Yes, the data of the intact bike frames provides information about the glueing technique for the carbon fiber frames. The defective bike frames contain information about weaknesses and unstable parts. Both information could be relevant for competitors.        \\
       
    \hline
    Collection        
        & Describe the data collection procedure and the data sources        & Data was systematically collected by our company during the first year as well as by our service partner  using  commercial industrial cameras of type [MODEL] under diffuse lighting conditions for both cases. We collected data by taking the images of the frames directly after production. Our service partner captured images of the frames it got from its customers. We merged the data and visually checked for homogeneity. After that we labeled the data using an external labeling company.  \\\cline{2-3}
        
        \pagebreak
        \cline{2-3}
        & Use data version control                                  & A data versioning system is applied in the cloud.\\\cline{2-3}
        & Consider prior requirements for data                      & The exact amount of data was collected as defined above.\\\cline{2-3}
        & Include and describe metadata                             & We include the type of bicycle frame as well as the exact date the image was taken. \\\cline{2-3}  
        & Involve domain experts and describe their involvement     & Domain experts were involved during the collection of requirements for the dataset. They also provided instructions and a gold standard dataset for the labeling company. They made several spot checks to control for correctness of the labels afterwards.       \\\cline{2-3}
        & Describe technical measures to ensure completeness of data & We performed an embedding followed by a clustering to identify groups of bike frames other than the different bike models. Those clusters were checked by domain experts. The numbers sold of our different bike models are accurately represented in the data. Also we tried to include enough defect data to make sure the trained model learns how a defect looks in the images.    \\\cline{2-3}
        & Describe and record edge cases                            & One edge case is where the use of to much glue could lead to a label defect. This also happened a few times during the labeling process.    \\\cline{2-3}
        & If personal data is used, make sure and document that all individuals know they are part of the data & No individuals are represented in this dataset.        \\\cline{2-3}
        & Describe if and how the data could be misused & It could be used by competitors to find weaknesses and important design choices in our product line.       \\\cline{2-3}
        & If fairness is identified as a risk, list sensitive attributes & No information is saved to reconstruct the identity of workers.  \\\cline{2-3}
        & If applicable, address data poisoning                     &  Not applicable.    \\\cline{2-3}
    
    \hline
    Labeling        
        & If applicable, describe the labeling process              
        &  The labeling company got a short briefing and a gold-standard labeling set by our domain experts. After the labeling we made spot checks on the dataset.        \\\cline{2-3}  
        & If applicable, describe how the label quality is checked  &  Domain experts perform spot checks. \\
            
    \hline
    Splitting        
        & Create and document meaningful splits with stratification                     & We divided the dataset into easy and hard data points. In the hard data points there were new bike models previously not seen by the model.\\\cline{2-3}  
        & Describe how data leakage is prevented                                        & We use the same camera type for all images. After training we perform heatmap analysis to ensure the model focuses on the bike frame.              \\\cline{2-3}  
        & Recommendation: test the splits and variance via cross-validation             & The variance between several evaluations on the test set is below 1\%.    \\\cline{2-3}  
            
        \pagebreak
        \cline{2-3}  
        & Recommendation: split dataset into difficult, trivial and moderate              & An additional (hard) test set was created that contains novel frames from the service provider that are not part of train data. This allows us to evaluate the applicability of our model to future products. \\\cline{2-3}  
        & Recommendation: put special focus on label quality of test data            & The gold-standard data is included in the test set and all test labeles were inspected by two separate experts. \\\cline{2-3}  
        & Reminder: Perform separate data preprocessing on the splits & Done. \\\cline{2-3}  
              
    \hline
    Preprocessing        
        & Document and motivate all processing steps that are a fixed part of the data   &  The image data is normalized from 0 to 1 to ensure a good behaviour of the convolutional neural network (CNN). \\\cline{2-3}  
        & Document whether the raw data can be accessed                                 &  The raw data is stored on our servers and can be accessed at any time by us. It is not publicly available.          \\\cline{2-3}  
        & If sensitive data is available highlight (pseudo)-anonymization               & No (pseudo)-anonymization needed. \\\cline{2-3}  
        & If fairness is a risk, highlight fairness specific preprocessing             & We do not record information to draw conclusion about the identity of specific workers.\\
             
    \hline
    Analyzing        
        & Understand and document characteristics of test and training data     &  We have 4\,700 intact bike frames and 500 bike frames with defects. 300 of the defective frames are from our own production and are identical to our frames, 200 are from the service provider. We use 20$\%$ of the data for the test split.\\\cline{2-3}   
        & Document why the data distribution fits the real conditions or why this is not necessary for the use case         &  Images were taken directly at our test production line and at the location of our service provider using real intact and defective frames.\\\cline{2-3}  
        & Document limitations such as errors, noise, bias or known confounders   &  It is possible that non-superficial defects cannot be identified by a visual check.\\\cline{2-3}  
    \hline
    Serving       
        & Describe how the dataset will be maintained in future &  New images from the production line and the service provider will be added to the dataset after labeling by the external company. \\\cline{2-3}  
        & Describe the storage concept   &   We operate a server with redundancy to store the image data.   \\\cline{2-3} 
        & Describe the backup procedure  &   A compressed and encrypted backup of the current dataset is saved at an external cloud storage provider.      \\\cline{2-3}  
        & If necessary, document measures against data poisoning                &  Not necessary.        \\
                      
    \bottomrule
\end{longtable}
}

\newpage
{\small
  \begin{longtable}{p{0.15\linewidth}|p{0.35\linewidth}p{0.5\linewidth}}
    \toprule
    \multicolumn{3}{c}{\bf{Model Card}}\\
    \toprule
    \bf{Step}   & \bf{Requirement}   &  \bf{Measures} \\
    \hline
    \hline
    General                     
        & Provide Name and details of a contact person 		    & Jane Doe, Project Manager\\\cline{2-3}
        & Provide Name of the person who created the model      & John Doe, ML Engineer\\\cline{2-3}
        & Document the creation date and version of the model   & 01-2023, Version 1.0 \\\cline{2-3}
        & Describe intended use of the model                    & Detection of defects in carbon fiber frames\\
        
    \specialrule{.075em}{0em}{0em}
    Description       
        & Describe the architecture of the used model    	& Custom CNN\\\cline{2-3}
        & Describe the used hyper-parameters         	    & Six convolution layers with ReLu activation followed by max-pooling. Sigmoid activation in the final layer. The model is trained for 200 epochs with learning rate 1e-6.\\\cline{2-3}
        & Document the training, validation, and test error     & Recall on train and both moderate and hard test set is 100\%. However, the precision is only around 90\%. Hence, the recall is good enough to apply the solution but an improved precision would further reduce  manual checks.\\\cline{2-3}
        & Document computation complexity, training time and energy consumption (for foundation models describe steps taken to reduce energy consumption) & Model was trained on cloud resources using 6 NVIDIA A100 GPUs for 32 hours. Estimated energy consumption: 70 kWh.\\
                        
    \specialrule{.075em}{0em}{0em}
    Explainability and interpretability    
        & Document the demand for explainability and interpretability 	& The model should provide heatmaps to workers to help them detect regions of interest.\\\cline{2-3}
        & Describe taken actions if any 			& GradCam (or similar) will be applied post-hoc during operation.\\   

    \specialrule{.075em}{0em}{0em}
    Feature engineering, feature selection and preprocessing   
        & Describe the consideration of explainability and interpretability (if relevant)  & Not relevant because there is no feature engineering.\\\cline{2-3}
        & Describe feature engineering and selection with (domain specific) reasoning   & No feature engineering.\\\cline{2-3}
        & Describe and reason of preprocessing steps  & Data augmentation in form of brightness and Gaussian noise.\\

    \specialrule{.075em}{0em}{0em}
    \pagebreak
    \specialrule{.075em}{0em}{0em}
    Model selection                     
        & Describe the consideration of explanibility, interpretability and fairness (if relevant) 	& A standard heatmap method must be applicable to the model.\\\cline{2-3}
        & Describe the base line model and its evaluation	& Multiple baselines were evaluated, a linear model, ViT and smaller and larger CNNs.\\\cline{2-3}
        & Describe the reasoning for the model choice 		& The final model was chosen based on a trade-off between performance and model size.\\\cline{2-3}
        & Describe why the complexity of model is justified and needed 	& Yes, smaller or simpler models performed worse. Larger models increased the inference time too much.\\\cline{2-3}
        & Document the comparison to other considered models 	  & Using a deeper model such as ResNet only marginally improves performance by less than 1\%.\\\cline{2-3}
        & Document the approach of hyper-parameter optimization      &  Random search. Increases test performance by around 5\%. \\\cline{2-3}
        & Describe the model evaluation                   &  The model is evaluated w.r.t. the recall and precision on both the moderate and hard test and, additionally, tested in production.\\
        
    \specialrule{.075em}{0em}{0em}
    Choice of metrics                   
        & Describe selected metrics and describe reasoning regarding use case and fairness    	& The evaluated metric is recall and an inference time below 0.5 seconds. Fairness is not applicable here.\\\cline{2-3}
        & Define the minimum requirement for use in production (domain specific reasons) 	& Recall above 99\% is the target. Either purely from the model or by including a human in the loop. A precision of 90\% is targeted but not mandatory. \\

    \specialrule{.075em}{0em}{0em}
    Model confidence                   
        & Document and quantify uncertainty of the model    & Model confidence was not assessed, since the model was trained w.r.t recall.\\\cline{2-3}
        & Document approach of dealing with uncertainty     & None, see above.\\

    \specialrule{.075em}{0em}{0em}
    Testing in real world setting       
        & Describe test design 		&  During the initial phase (expected to be the first year) all frames will additionally be checked manually. During this phase, faulty frames will be injected into the line in order to validate the AI solution.\\\cline{2-3}
        & Describe possible risks, edge cases and worst case scenarios and create (or simulate) them if possible & Edge cases are tested both using the different test sets (hard and medium) and through injection of faulty parts.\\\cline{2-3}
        & Describe limitations and shortcomings of the model & Change of production process or architecture of frames might introduce new faults, that might not be detcted by the model.\\\cline{2-3}
        & Describe the test results		& The test is still running but revealed problems of the AI solution under specific lighting conditions\\\cline{2-3}
        & Explain the derived actions 	& The dataset was extended to include such conditions\\
    \bottomrule
  \end{longtable}
}

\newpage
{\small
\renewcommand\arraystretch{1.3}
\begin{longtable}{p{0.15\linewidth}|p{0.35\linewidth}p{0.5\linewidth}}
    \toprule
    \multicolumn{3}{c}{\bf{Operation Card}}\\
    \toprule
    \bf{Step}   & \bf{Requirement}   &  \bf{Measures} \\
    \hline
    \hline
    \multirow{7}{=}{General} 
    Scope and aim of monitoring     
        & Describe monitored components & The CV system as a whole is monitored. Input and output of the model are evaluated separately. Additionally the function of the cameras is monitored.\\\cline{2-3}
        & Assess risks and potential dangers according to Use case card & Undetected faults lead to the shipment of broken frames. False Alarms might lead to ineffective manual inspections and trust issues of personnel.\\\cline{2-3}
        & List safety measures for risks & Monitoring of input and output data. Regularly reoccurring computed tomography enables assessment of model performance. Manual inspections enable the assessment of precision.\\\cline{2-3}
                                    
    \hline
    Operating concept               
        & Create and document utilisation concept & If the CV system reports possible damage of a frame, staff does a standardized thorough visual and tactile examination.\\\cline{2-3}
        & Plan staff training & Each employee who conducts the examination needs to take a training session with test. Staff training of the user interface and AI basics is mandatory. \\\cline{2-3}
        & Determine responsibilities & The shift supervisor is responsible for the examination and dedicated IT  personnel for the CV system.\\\cline{2-3}
        
    \hline
    Autonomy of the application     
        & Document decision-making power of AI  & The AI system checks autonomously and only notifies personnel if faults are detected. The user interface allows checking the system and the images made by the cameras at all times. \\\cline{2-3}
        & Determine process to overrule decisions of the AI & Regularly reoccurring computed tomography tests overrule the CV systems decision. Personnel can overrule the AI system's decision after manual inspection of a frame.\\\cline{2-3}
    
    \hline
    Responsibilities and measures   
        & Component wise: assessed risk, control interval, responsibility, measures for emergency & Risks of system components (CV model, cameras, data processing pipeline and user interface) are assessed separately. Further detailed information is attached to this document.\\\cline{2-3}
    
    \hline
    \pagebreak
    \hline
    Model performance               
        & Monitor input and output & Input: Distribution of pixel values are monitored. Output: Number and distribution of defective frames is monitored over a fixed time period.\\\cline{2-3}
        & Detect drifts in input data & Input data is monitored with torch drifts Kernel MMD drift detector method.\\\cline{2-3}
        & Document metric in use to monitor model performance & Model performance is monitored via precision and recall. If precision is over 95\% or under 85\% personnel will be alerted. If the CV system  misses a defective frame (detected by computed tomography) personnel will be alerted immediately. If the percentage of defective frames increases, or percentage of defective frames decreases over time, personnel is alerted immediately.\\\cline{2-3}
    
    \hline
    AI Interface                    
        & Establish transparent decision making process & Decision making is made transparent with heatmaps and displayed on the user interface. \\\cline{2-3}
        & Establish insight in model performance on different levels & Production staff: current frame and heatmaps displayed. Display of heatmaps of frames identified as defective.
        IT personnel: Camera images and heatmaps analogous to production staff. Additionally, access to logs, performance metrics and monitoring as described.\\\cline{2-3}
    
    \hline
    IT security                     
        & Document individual access to server rooms & Access to server rooms is restricted to authorized personnel only\\\cline{2-3}
        & Set and document needed clearance level for changes to AI/Deployment/Access regulation & Only trained and authorized personnel can make changes to the CV system. Changes need to be verified by a second person to be applied.\\\cline{2-3}
        & Establish and audit ISMS & IT system in which the AI application will be integrated is certified according to ISO 270001. A review of the system will be conducted with an IT-security manager to evaluate the need for re-certification.\\ \cline{2-3}
    
    \hline
    Privacy                         
        & Justify and document use or waiver of a privacy preserving algorithm & Not applicable since no personal data is collected.\\\cline{2-3}
        & If applicable, document privacy algorithm and due changes in the monitoring of output data & Not applicable.\\\cline{2-3}
    
    \hline
    MLOps                           
        & Establish versioned code repository of AI, training and deployment & Version control is implemented with Git and MLflow.\\\cline{2-3}
        & Establish maintenance and update schedule & The ML model is retrained every 3 months (if new data is available) on a blend of old and new data or if a drift is detected.\\\cline{2-3}
        & Set regulation for the retraining of the AI and decision basis for the replacement of a model & The model is updated with new training data. The update goes live, if the recall does not drop and precision increases or remains constant. If performance measures drop twice in a row after retraining, data will be analyzed.\\\cline{2-3}
    
    \hline
    Testing and rollout             
        & Determine responsibilities for updates & Updates are managed by dedicated IT personnel. Changes need to be verified by a second person to be applied.\\\cline{2-3}
        & Document software tests & Software tests are automated and documented.\\\cline{2-3}
        & Set period of time that an update must function stably before it is transferred to live status & New software and model versions run separately on the test environment parallel to the production environment for one week. Performance is evaluated and compared to the old version before deployment.\\
                                    
    \bottomrule
\end{longtable}
}

\newpage
\section{Content Themes to Describe a Dataset according to \citet{Gebru2021_DataSheets}}
\label{app:datasheets_content_themes}
(1) About the publishers of the dataset and access to them \\
(2) The funding of the dataset \\
(3) The access restrictions and policies of the dataset \\
(4) The wipeout and retention policies of the dataset \\
(5) The updates, versions, refreshes, additions to the data of the dataset\\
(6) Detailed breakdowns of features of the dataset \\
(7) If there are attributes missing from the dataset or the dataset’s documentation\\
(8) The original upstream sources of the data \\
(9) The nature (data modality, domain, format, etc.) of the dataset\\
(10) What typical and outlier examples in the dataset look like\\
(11) Explanations and motivations for creating the dataset \\
(12) The intended applications of the dataset \\
(13) The safety of using the dataset in practice (risks, limitations, and trade-offs)\\
(14) The maintenance status and version of the dataset \\
(15) Difference across previous and current versions of the dataset\\
(16) Expectations around using the dataset with other datasets or tables (feature engineering, joining, etc.)\\
(17) The data collection process (inclusion, exclusion, filtering criteria)\\
(18) How the data cleaned, parsed, and processed (sampling, filtering, etc.)\\
(19) How the data was rated in the dataset, its process, description and/or impact\\
(20) How the data was labelled in the dataset, its process, description and/or impact\\
(21) How the data was validated in the dataset, its process, description and/or impact\\
(22) The past usage and associated performance of the dataset (eg. models trained)\\
(23) Adjudication policies related to the dataset (labeller instructions, inter-rater policies, etc.\\
(24) Regulatory or compliance policies associated with the dataset (GDPR, licensing, etc.)\\
(25) Dataset Infrastructure and/or pipeline implementation\\
(26) The descriptive statistics of the dataset (mean, standard deviations, etc.)\\
(27) Any known patterns (correlations, biases, skews) within the dataset\\
(28) Any socio-cultural, geopolitical, or economic representation of people in the dataset\\
(29) Fairness-related evaluations and considerations of the dataset\\
(30) Definitions and explanations for technical terms used in the dataset’s documentation (metrics, industry-specific terms, acronyms)\\
(31) Domain-specific knowledge required to use the dataset\\

\newpage
\bibliography{sample-base}

\end{document}